\documentclass[sigconf]{acmart}
\usepackage{siunitx}
\usepackage{multirow}
\usepackage{pifont}
\usepackage{cleveref}
\usepackage{titlesec}
\usepackage{enumitem}
\usepackage{balance}
\titlespacing*{\paragraph}{0em}{*1}{*1}

\AtBeginDocument{%
  }

\copyrightyear{2025}
\acmYear{2025}
\setcopyright{cc}
\setcctype{by}
\acmConference[MM '25]{Proceedings of the 33rd ACM International Conference on Multimedia}{October 27--31, 2025}{Dublin, Ireland}
\acmBooktitle{Proceedings of the 33rd ACM International Conference on Multimedia (MM '25), October 27--31, 2025, Dublin, Ireland}
\acmDOI{10.1145/3746027.3754769}
\acmISBN{979-8-4007-2035-2/2025/10}

\acmSubmissionID{532}

\begin{document}

\title{Occlusion-Aware Temporally Consistent Amodal Completion for 3D Human-Object Interaction Reconstruction}
% \title{Occlusion-Aware and Consistent Amodal Completion for 3D Human-Object Interaction Reconstruction}


\author{Hyungjun Doh}
\authornote{Three authors contributed equally to this research.}
\orcid{0009-0008-3154-1201}
\affiliation{%
  \institution{Elmore Family School of Electrical
and Computer Engineering}
  \institution{Purdue University}
  \streetaddress{585 Purdue Mall}
  \city{West Lafayette}
  \state{IN}
  \country{USA}
  \postcode{47907}
}
\email{hdoh@purdue.edu}

\author{Dong In Lee}
\authornotemark[1]
\authornote{Work done at Purdue University while a visiting scholar.}
\orcid{0000-0001-5426-6216}
% \affiliation{%
%   \institution{Elmore Family School of Electrical
% and Computer Engineering}
%   \institution{Purdue University}
%   \streetaddress{585 Purdue Mall}
%   \city{West Lafayette}
%   \state{IN}
%   \postcode{47907}
% }
\affiliation{
  \institution{Department of Artificial Intelligence}
  \institution{Korea University}
  \streetaddress{}
  \city{Seoul}
  \state{}
  \country{Republic of Korea}
  \postcode{}
}
\email{dilee99@korea.ac.kr}

\author{Seunggeun Chi}
\authornotemark[1]
\orcid{0000-0001-6965-6938}
\affiliation{%
  \institution{Elmore Family School of Electrical
and Computer Engineering}
  \institution{Purdue University}
  \streetaddress{585 Purdue Mall}
  \city{West Lafayette}
  \state{IN}
  \country{USA}
  \postcode{47907}
}
\email{chi65@purdue.edu}

\author{Pin-Hao Huang}
\orcid{0009-0002-1545-0058}
\affiliation{%
  \institution{Honda Research Institute USA}
  \city{San Jose}
  \state{CA}
  \country{USA}
  \postcode{47907}
}
\email{pin-hao_huang@honda-ri.com}

\author{Kwonjoon Lee}
\orcid{0000-0002-1433-551X}
\affiliation{%
  \institution{Honda Research Institute USA}
  \city{San Jose}
  \state{CA}
  \country{USA}
  \postcode{47907}
}
\email{kwonjoon_lee@honda-ri.com}

\author{Sangpil Kim}
\orcid{0000-0001-8639-5135}
\affiliation{
  \institution{Department of Artificial Intelligence}
  \institution{Korea University}
  \streetaddress{}
  \city{Seoul}
  \state{}
  \country{Republic of Korea}
  \postcode{}
}
\email{spk7@korea.ac.kr}

\author{Karthik Ramani}
\orcid{0000-0001-8639-5135}
\affiliation{%
  \institution{Elmore Family School of Electrical
and Computer Engineering}
  \institution{School of Mechanical Engineering}
  \institution{Purdue University}
  \streetaddress{585 Purdue Mall}
  \city{West Lafayette}
  \state{IN}
  \country{USA}
  \postcode{47907}
}
\email{ramani@purdue.edu}

\renewcommand{\shortauthors}{Hyungjun Doh et al.}


\begin{abstract}
We introduce a novel framework for reconstructing dynamic human–object interactions from monocular video that overcomes challenges associated with occlusions and temporal inconsistencies. Traditional 3D reconstruction methods typically assume static objects or full visibility of dynamic subjects, leading to degraded performance when these assumptions are violated—particularly in scenarios where mutual occlusions occur. To address this, our framework leverages amodal completion to infer the complete structure of partially obscured regions. Unlike conventional approaches that operate on individual frames, our method integrates temporal context, enforcing coherence across video sequences to incrementally refine and stabilize reconstructions. This template-free strategy adapts to varying conditions without relying on predefined models, significantly enhancing the recovery of intricate details in dynamic scenes. We validate our approach using 3D Gaussian Splatting on challenging monocular videos, demonstrating superior precision in handling occlusions and maintaining temporal stability compared to existing techniques. Project Page: \href{https://danieldoh.github.io/OTA-3DHOI/}{\textcolor{blue}{https://danieldoh.github.io/OTA-3DHOI/}}

\end{abstract}

\begin{CCSXML}
<ccs2012>
   <concept>
       <concept_id>10010147.10010178.10010224.10010245.10010249</concept_id>
       <concept_desc>Computing methodologies~Shape inference</concept_desc>
       <concept_significance>500</concept_significance>
       </concept>
   <concept>
       <concept_id>10010147.10010178.10010224.10010245.10010254</concept_id>
       <concept_desc>Computing methodologies~Reconstruction</concept_desc>
       <concept_significance>100</concept_significance>
       </concept>

 </ccs2012>
\end{CCSXML}

\ccsdesc[500]{Computing methodologies~Shape inference}
\ccsdesc[100]{Computing methodologies~Reconstruction}

\keywords{Human-Object Interaction, Amodal Completion, Temporal Consistency}

\begin{teaserfigure}
  \centering
  \includegraphics[width=0.98\textwidth]{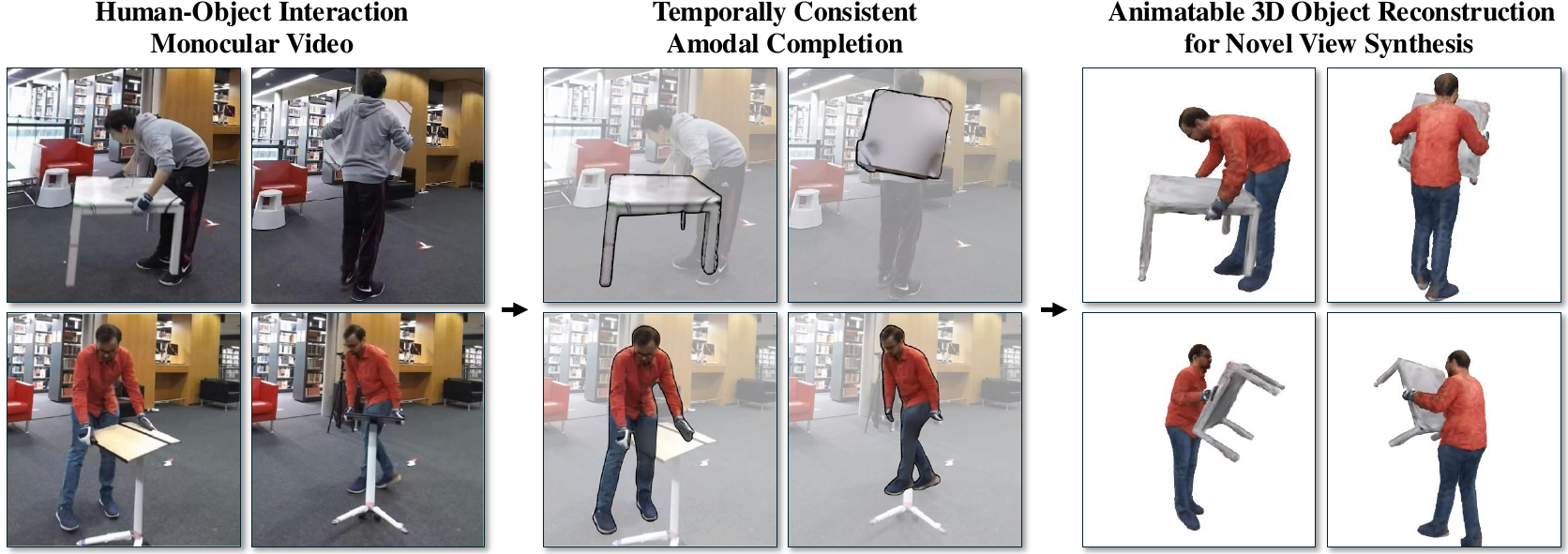}
  \vspace{-1.0em}
  \caption{(\textit{left}) In human–object interaction (HOI) scenarios, occlusions frequently affect both the human and the object. (\textit{middle}) We inpaint the occluded regions while preserving temporal consistency for both entities across frames. (\textit{right}) Leveraging the temporally consistent image sequences, we reconstruct the human and object using a 3D Gaussian splatting representation, enabling animatable 3D HOI applications.}
  \label{fig:teaser}
  \vspace{-0.5em}
\end{teaserfigure}

\maketitle



% \newcommand{\kwon}[1]{\textcolor{blue}{[Kwonjoon: #1 ]}}

\vspace{-1.em}
\section{Introduction} 
\label{sec:intro}

Understanding the complex interplay between humans and objects is a fundamental challenge in computer vision and robotics, with broad implications for both theory and real-world applications. This understanding is critical for enabling technologies such as autonomous robots that navigate complex human environments and augmented or virtual reality systems that rely on precise 3D reconstructions. Despite significant progress, reliably interpreting human-object interactions (HOI) in dynamic, real-world settings remains an open problem. Visual complexities—especially mutual occlusions—often obscure critical details, underscoring the need for enhanced 3D reconstruction techniques that capture and animate these interactions effectively.

Multi-view 3D reconstruction~\cite{kerbl3Dgaussians, mildenhall2021nerf, schonberger2016structure} methods have proven effective in generating detailed models from multiple images. Object reconstruction techniques~\cite{yang2024gaussianobject,jain2021putting, sun2022direct, zhu2024fsgs} are generally designed for static scenes, where the object remains stationary. In contrast, human reconstruction methods~\cite{jiang2022neuman, kocabas2024hugs, hu2024gaussianavatar} typically handle dynamic subjects assuming full visibility. However, in HOI scenarios, these assumptions break down. Both humans and objects are in motion, leading to frequent and mutual occlusions as illustrated in~\Cref{fig:teaser}, resulting in ambiguous visual cues that complicate the reconstruction process. The presence of dynamic and mutual occlusions introduces complexities beyond the scope of static object or dynamic full-visibility human reconstruction methods.

Amodal completion~\cite{zhou2023amodal, chen2023amodal3d, ozguroglu2024pix2gestalt, li2022compositional, nguyen2022disentangled, wu2022selfsupervised, xu2024amodal, chi2025contactawareamodalcompletionhumanobject} has emerged as a promising strategy to mitigate these occlusion challenges by inferring the complete structure of partially hidden regions. However, most existing approaches are designed for static scenes. Although they perform well in controlled environments, they often falter in dynamic, real-world settings. A key limitation is their reliance on image-level completion, which overlooks the rich temporal context provided by adjacent sequences in HOI cases. Without leveraging temporal information, reconstructions derived from amodal-completed images can suffer from inconsistencies, ultimately compromising the overall quality and stability of the models.

Recent research~\cite{teed2020raft, wang2024sea, zhou2024upscale, zhang2024avid, yang2023rerender, cong2023flatten, jeong2022imposing} suggests that incorporating temporal consistency can effectively address these challenges. By enforcing coherence across consecutive video frames, models can integrate information from multiple viewpoints more effectively, leading to more accurate recovery of occluded regions. This strategy also enables the incremental refinement of hidden details over time, yielding reconstructions that more faithfully capture the dynamic nature of human–object interactions.

Building on these insights, we propose a novel, temporally consistent amodal completion framework tailored for monocular video input. By integrating temporal context into the amodal completion process, our approach overcomes the limitations of traditional frame-by-frame methods, significantly enhancing both stability and realism. 
Specifically, our framework consists of three key components. 
First, we introduce \textit{Bidirectional Temporal Feature (BTF) Warping}, which leverages optical flow to warp latent features from both past and future frames into the current frame’s latent space, enabling effective propagation of temporal cues and aligns spatially meaningful information across time. 
Second, we present a \textit{Temporal Fusion Attention} mechanism that adaptively aggregates these temporally aligned features, producing a coherent and semantically enriched latent representation.
Third, we propose a \textit{Template-free Occlusion Identification} strategy that combines 2D segmentation and 3D projections to localize occlusions without relying on predefined templates. Finally, we apply amodal completion, guided by the temporal-aware latent features and precise occlusion masks, to complete missing regions with high fidelity.
This unified pipeline enables robust recovery of occluded structures across time, ensuring temporal consistency and accurate amodal completion. 
Unlike prior approaches, our method generalizes effectively to dynamic, real-world human–object interaction scenarios.

We validate the effectiveness of our method through extensive experiments on two public datasets, BEHAVE~\cite{bhatnagar2022behave} and InterCap~\cite{huang2022intercap, huang2024intercap}, which present severe occlusions and diverse human–object interactions. Our framework demonstrates consistently superior performance in both quantitative metrics and qualitative comparisons.
Moreover, we extend our evaluation to reconstructing 3D human–object interactions from monocular video sequences using 3D Gaussian Splatting~\cite{kerbl3Dgaussians} (3DGS), enabling various 3D application of animatable HOI including novel-view synthesis.% and novel pose synthesis.

In summary, our contributions include: 
\vspace{-1.em}
\begin{itemize} 
\item \textbf{Temporally Consistent Completion Method:} We carefully design a Bidirectional Temporal Feature (BTF) Warping and a Temporal Fusion Attention mechanism that incorporate temporal context to amodal completion by adaptive fusion of aligned features across past and future frames.

\item \textbf{Template-free Occlusion Identification Strategy:} We propose a hybrid approach that integrates 2D segmentation with 3D point cloud projection to accurately localize occluded regions. This method introduces a novel masking technique that identifies occlusions efficiently, without the need for predefined templates.

\item \textbf{Application to 3D Reconstruction:} To the best of our knowledge, our work is the first approach to reconstruct photo-realistic and animatable 3D human-object interactions from monocular videos. By leveraging 3D Gaussian Splatting~\cite{kerbl3Dgaussians}, we demonstrate that our pipeline is applicable to various subtasks, including novel-view and novel-pose synthesis for HOI.
\vspace{-0.5em}
\end{itemize}

\section{Related Work}
\label{sec:related_work}
%-------------------------------------------------------------------------

\vspace{-0.3em}
\subsection{3D Reconstruction}
3D reconstruction~\cite{kerbl3Dgaussians,mildenhall2021nerf,schonberger2016structure} from multi-view images has demonstrated versatility across various fields. In particular, 3D reconstruction using explicit representations~\cite{kerbl3Dgaussians} has emerged as an effective method for capturing 3D information. This representation has been successfully employed in 3D scene reconstruction~\cite{Fu_2024_CVPR,scaffoldgs,lin2024vastgaussian}, object reconstruction~\cite{yang2024gaussianobject,jain2021putting, sun2022direct, zhu2024fsgs}, and human reconstruction~\cite{jiang2022neuman, kocabas2024hugs, hu2024gaussianavatar}. However, these methods often face challenges when occlusions occur. To mitigate the occlusion problem in human reconstruction, few research proposed~\cite{lee2024guess, sun2024occfusion} a diffusion-based method. However, these methods implicitly secure temporal consistency through a 2D diffusion prior, not guaranteeing temporal consistency.
In contrast to existing approaches that primarily address either human or static objects, our method specifically targets cases involving human-object interactions via template-free amodal completion.

\vspace{-0.7em}
\subsection{Amodal Completion}
Amodal completion focuses on inferring the hidden shapes and appearances of occluded objects. Recently, various works leverage generative models to reconstruct missing regions. For instance, Chi et al.~\cite{chi2025contactawareamodalcompletionhumanobject} completes occlusion using contact points while pix2gestalt~\cite{ozguroglu2024pix2gestalt} synthesizes “wholes” from partial observations and Xu et al.~\cite{xu2024amodal} expands occluded areas with context-awareness. Other studies propose transformer-based amodal segmentation~\cite{zhou2023amodal}, leverage 3D shape priors to guide amodal masks~\cite{chen2023amodal3d}, or learn compositional scene representations to handle complex occluders~\cite{li2022compositional, nguyen2022disentangled}. Self-supervised frameworks alsowemerged for amodal completion by aligning visible parts with plausible hidden geometry~\cite{wu2022selfsupervised, kim2023monocular}. However, these single-image methods often struggle with dynamic human-object interactions, where severe occlusions may shift from frame to frame—leading to temporally inconsistent reconstructions. As an alternatives, video inpainting methods~\cite{lu2023vdt} has also been introduced, however these methods are usually for removing subjects well fitted to background and environment, showing limitation on HOI cases. Our approach addresses these issues by integrating temporal cues, ensuring stable amodal completions across sequences.

\vspace{-0.7em}
\subsection{Temporal Consistency}
Ensuring temporal consistency across image sequences is essential for stable and coherent video processing. Traditional approaches often rely on optical flow to propagate features between frames~\cite{teed2020raft, wang2024sea, jeong2022imposing}. Recent advancements in video diffusion have incorporated explicit temporal modeling to improve alignment and consistency across frames.
Several state-of-the-art techniques~\cite{zhou2024upscale, zhang2024avid, yang2023rerender, cong2023flatten} adopt flow-guided recurrent architectures, enabling temporally aware representations for applications such as video super-resolution, inpainting, video-to-video translation, and text-driven video editing. Complementary strategies include transformer-based frame propagation~\cite{zhou2023propainter} and cross-frame feature matching through diffusion-based tokens~\cite{geyer2023tokenflow}, both of which contribute to enhanced coherence in complex temporal dynamics.
While these methods primarily focus on video generation and restoration tasks, our approach uniquely leverages optical flow and feature warping mechanisms to specifically address occlusions and ensure temporally consistent reconstructions in 3D HOI scenarios.

\begin{figure*}[t]
    \centering
    % \fbox{\makebox[0.95\textwidth][c]{\rule{0pt}{5cm} Placeholder for Overall Pipeline}}
    \includegraphics[width=0.95\textwidth]{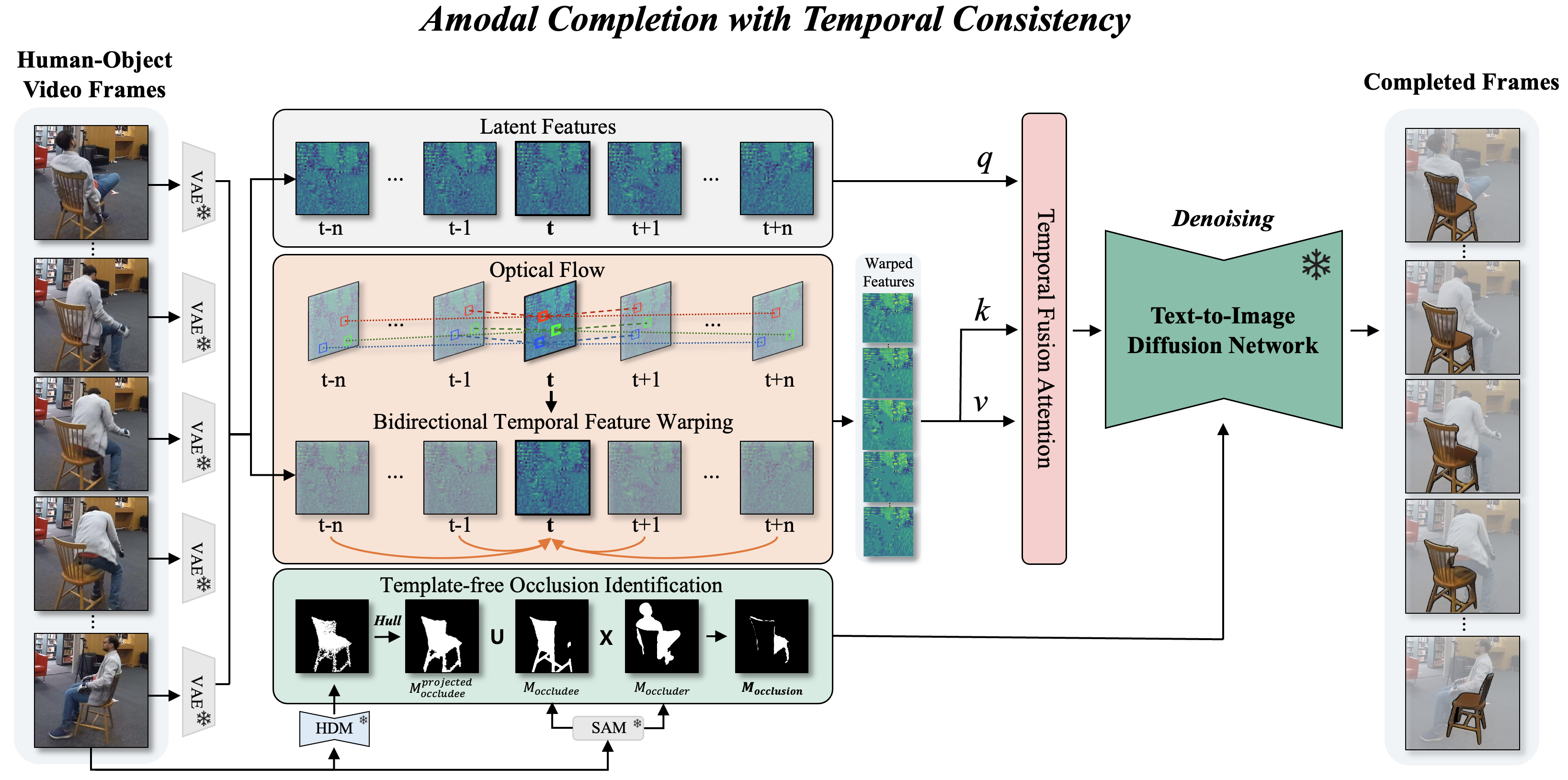}
    \vspace{-1.5em}
    \caption{Overview of Our Framework: Given a human–object interaction (HOI) monocular video, our framework performs amodal completion through (1) Bidirectional Temporal Feature Warping via optical flow, (2) Temporal Fusion Attention for aggregating multi-frame context, (3) Template-free Occlusion Identification using 2D and 3D cues, and (4) temporally-aware amodal completion. This design enables temporally consistent and accurate amodal completion in complex HOI scenarios.}
    \label{fig:pipeline}
    \vspace{-1.5em}
\end{figure*}

\vspace{-0.5em}
\section{Method}
\label{sec:method}

Our goal is to achieve temporal consistency while completely recovering occluded regions for accurate 3D reconstruction of Human–Object Interactions (HOI) from monocular video. This is challenging due to frequent and severe occlusions in HOI scenarios. To address this challenge, we propose a novel pipeline that integrates temporal context into the amodal completion process.

We begin by formulating the temporally consistent amodal completion problem, clearly defining the inputs, outputs, and objectives in~\Cref{sec:problem}. Based on this formulation, our method consists of the following four main components, as illustrated in~\Cref{fig:pipeline}. First, in~\Cref{sec:BTF}, we introduce Bidirectional Temporal Feature (BTF) warping, which aligns features across neighboring frames using optical flow and latent feature warping. Second, in~\Cref{sec:TFA}, we present Temporal Fusion Attention, a mechanism that dynamically fuses the temporally aligned features into a coherent representation. Third, in~\Cref{sec:occ}, we describe a template-free occlusion identification strategy that accurately localizes occluded regions. The last component, explained in ~\Cref{sec:amodal}, is temporally-aware amodal completion mechanism. Consequently, our method enables reliable 3D reconstruction through temporally consistent amodal completion, as detailed in the supplementary materials.

\vspace{-0.7em}
\subsection{Problem Definition}\label{sec:problem}
We address the problem of filling in occluded regions in a sequence of video frames such that the completed regions are both semantically plausible and temporally coherent. For each frame, let \( I_{in} \in \mathbb{R}^{H \times W \times 3} \) be the segmented image containing only the visible regions, and let \( M_{in} \in \{0,1\}^{H \times W}\) be a binary mask, where 1 indicates the occluded regions to be inpainted. In addition, we provide a temporally-aware latent feature \( z_{in} \), aggregated from temporally aligned features across neighboring frames, and a text prompt \( P \) as textual guidance. Adapting the problem definition of~\cite{xu2024amodal}, we formalize the inpainting process with the diffusion model~\cite{rombach2022high} guided by both latent and mask inputs as:
\begin{equation}
    I_{out} = \mathbf{F}_{s \to e}\left(I_{in},\, M_{in},\, z_{in},\, P\right),
\end{equation}
where \(\mathbf{F}_{s \to e}\) denotes the diffusion process from denoising time step \( s \) to \( e \). The effectiveness of our framework critically depends on two factors: (1) the quality of the latent feature \( z_{in} \), which provides temporally consistent and semantically rich context, and (2) the accuracy of the occlusion mask \( M_{in} \), which precisely localizes the region to be completed. In the following sections, we describe how each component is constructed to ensure effective amodal completion under complex human–object interaction scenarios.

\vspace{-0.7em}
\subsection{Bidirectional Temporal Feature Warping} \label{sec:BTF}

To ensure temporal consistency across consecutive frames, we introduce a Bidirectional Temporal Feature (BTF) warping method that aligns latent features from both past and future frames to the current frame. As part of this approach, we define a temporal support set \(\mathcal{S}(t)\) at each time step \(t\), consisting of past and future neighboring frames:
\vspace{-0.5em}
\begin{equation}
\mathcal{S}(t) = \{t\!-\!n,\dots, t\!-\!1, t\!+\!1, \dots,t\!+\!n\}.
\end{equation}
We empirically choose $n = 7$ aggregating total 14 neighboring features. For each element in the support set \(\tau \in \mathcal{S}(t)\), we estimate optical flow from frame \(I^\tau\) to the reference frame \(I^t\) to spatially align their latent representations. The optical flow is computed as:
\begin{equation}
o^t_{\tau} = \mathrm{Flow}(I^\tau, I^t), \quad \forall \, \tau \in \mathcal{S}(t),
\end{equation}
where \(\mathrm{Flow}(\cdot, \cdot)\) denotes an off-the-shelf optical flow estimation network~\cite{wang2024sea}. 

We then encode latent features using a pretrained Variational Autoencoder (VAE)~\cite{kingma2013auto} \(\mathcal{E}\) as utilized in diffusion models~\cite{rombach2022high}, defined as \(z^{\tau} = \mathcal{E}(I^{\tau})\). To spatially align the latent representations across frames, the encoded feature \(z^{\tau}\) is subsequently warped into the coordinate space of frame \(t\) using the estimated optical flow \(o^t_{\tau}\) which is bilinearly scaled to match the resolution of \(z^{\tau}\):
\begin{equation}
\hat{z}^{t}_{\tau} = \mathrm{Warp}(z^\tau, o^{t}_{\tau}), \quad \forall \, \tau \in \mathcal{S}(t).
\end{equation}

The warped features \(\hat{z}^{t}_{\tau}\) serve as temporally aligned observations centered at frame \(I^t\), enabling the aggregation of multi-frame context from both past and future frames. By referencing complementary information from adjacent frames, our framework can recover regions that are occluded or missing in the current view in complex human-object interaction scenarios.

% -------------------------------------------------------------------%
\vspace{-0.5em}
% \paragraph{\textbf{Temporal Fusion Attention}}
\subsection{Temporal Fusion Attention} \label{sec:TFA}

While the warped features are temporally aligned, it is crucial to effectively incorporate temporal context from each warped feature into a unified latent representation. Therefore, we propose a Temporal Fusion Attention mechanism that selectively integrates complementary cues from neighboring frames. Specifically, we employ a cross-attention strategy to aggregate temporally aligned latent features within the support set \(\mathcal{S}(t)\). At each time step \(t\), the queries, keys, and values are defined as:
\begin{equation}
Q^t = z^{t}, \quad K^t = V^t = \text{concat}(\hat{z}_{t-n}^t,\cdots,\hat{z}_{t+n}^t),
\label{eq:qkv}
\end{equation}
where \(Q^t\in \mathbb{R}^{1 \times d}\) represents the latent feature from the current frame, and \(K^t\in \mathbb{R}^{2n \times d}\), \(V^t\in \mathbb{R}^{2n \times d}\) represent the warped latent features aligned from adjacent frames. The attention-weighted representation is computed via scaled dot-product attention:
\begin{equation}
z_{fused}^t = \mathrm{Attn}(Q^t, K^t, V^t) = \mathrm{Softmax}\!\left(\frac{Q^t \cdot K^t}{\sqrt{d}}\right)V^t,
\end{equation}
where \(d\) denotes the feature dimensionality. By leveraging temporally warped features from both past and future frames, this mechanism selectively integrates complementary information that is not visible in the current view, thereby improving occlusion completion. This yields \(z_{fused}^t\), a semantically rich context that captures occluded structures and interaction context from neighboring frames. By emphasizing temporally relevant and contextually informative features, this attention mechanism significantly enhances coherence and robustness across frames, thereby facilitating high-quality and stable amodal completion.

% \vspace{-1.em}
\vspace{-0.5em}
\subsection{Template-free Occlusion Identification} \label{sec:occ}

To accurately localize the occluded area, it is essential to estimate the complete shape of the target—referred to as the occludee—which is the object of interest partially hidden in the scene. Therefore, We employ the Hierarchical Diffusion Model (HDM)~\cite{xie2023template_free}, a template-free approach specifically designed for reconstructing human-object interactions. HDM facilitates the inference of a comprehensive 3D point cloud representing the complete geometry of human and object. We then project this 3D point cloud onto the 2D image plane using a projection function \(\pi: \mathbb{R}^3 \to \mathbb{R}^2\). The resulting set of projected points is denoted as
$C = \{ \mathbf{u}_i \in \mathbb{R}^2 \mid \mathbf{u}_i = \pi(\mathbf{U}_i), \, \mathbf{U}_i \in \mathbb{R}^3, \, i = 1, 2, \dots, n \}$,
where each \(\mathbf{u}_i = (x_i, y_i)\) corresponds to the 2D coordinates of the projection of the 3D point \(\mathbf{U}_i\). This set \(C\) thus represents the 2D image plane positions of all projected points from the original 3D geometry. To extract a coherent shape from these points, we apply a concave hull algorithm~\cite{bellockk_alphashape} that constructs a tight boundary around the set. Unlike a convex hull, which encloses the outermost points with the smallest convex polygon, the concave hull denoted as $\text{ConcaveHull(}\!\cdot\!\text{)}$ allows for inward indentations, enabling the capture of non-convexities and finer geometric details. This leads to a more accurate approximation of the object's shape. The projected full shape mask of the target is then defined as:
\vspace{1pt}
\begin{equation}
M^{projected}_{occludee} = \text{ConcaveHull}(C).
\end{equation}
To ensure a more robust  estimation of the target's shape, we construct a fused mask, \(M_{union}\), by combining the segmented visible mask \(M^{visible}_{occludee}\), obtained by SAM2~\cite{ravi2024sam2}, with the projected full shape mask \(M^{projected}_{occludee}\):
\begin{equation}
M_{union} = M^{visible}_{occludee} \cup M^{projected}_{occludee},
\label{eq:fused_mask}
\end{equation}
which allows the two masks to effectively complement one another, integrating both observed and inferred regions of the target. The  Figure~\ref{fig:pipeline} demonstrates how the visible and projected masks are combined to yield a more complete representation.

Finally, we introduce an occluder mask \(M_{occluder}\), also segmented using SAM2, representing the region that occludes the target. The occluder can be either a human or an object, depending on the target of interest. The actual occlusion mask \(M_{occlusion}\), which delineates the occluded area of the target, is determined by intersecting the fused mask \(M_{union}\) with the occluder mask \(M_{occluder}\):
\vspace{-0.3em}
\begin{equation}
M_{occlusion} = M_{union} \cap M_{occluder}.
\label{eq:occlusion_mask}
\vspace{-0.3em}
\end{equation}
With this approach, we precisely identify the occluded region of the target object without relying on any predefined templates, significantly enhancing the generalizability and robustness.

% -------------------------------------------------------------------%

% \paragraph{\textbf{Amodal Completion}} 
\vspace{-0.7em}
\subsection{Temporally-aware Amodal Completion} \label{sec:amodal}
With temporally fused features \(z_{fused}^t\) and the occlusion regions \(M^t_{occlusion}\) identified in previous stages, we inpaint the occluded region of the target using the diffusion inpainting pipeline. To achieve temporally consistent amodal completion, we first apply occludee mask \(M^t_{occludee}\) to the input image \(I\) thereby removing background distractions and encouraging the model to focus on the target:\vspace{-0.5em}
\begin{equation}
I_{occludee}^t = M_{occludee}^t \odot I^t.
\label{eq:maksing}
\end{equation}
Next, we use the occlusion mask \(M^t_{occlusion}\) to specify the region that should be completed. To incorporate temporal context, we leverage a temporally-aware latent feature \(z^t_{fused}\), applying mask \(\hat{M}^t_{union}\) to ensure that latent representation only affects the target. Note that \(\hat{M}^t_{union}\) is bilinearly downsampled mask to match the dimension of \(z^t_{fused}\). A corresponding text prompt \(P\) is also provided to guide the inpainting process. This process can be written as:
\begin{equation}
I_{out}^t = {\mathbf{F}}_{s \to e}(I_{occludee}^t,\, M_{occlusion}^t,\, z_{fused}^t \odot \hat{M}_{union}^t,\, P).
\end{equation}
Through the integration of temporal cues, the diffusion inpainting pipeline reliably fills in the occluded regions while maintaining temporal consistency across frames. Consequently, our framework effectively resolves ambiguities due to occlusions and complex dynamics inherent in realistic human–object interactions, improving the quality and stability of amodal completion outcomes.

\begin{table*}
    \centering
    \renewcommand{\arraystretch}{1.3}
    \begin{tabular}{l S[table-format=2.2] S[table-format=2.2] S[table-format=3.2] S[table-format=2.2] S[table-format=2.2] S[table-format=2.2] S[table-format=2.2] S[table-format=3.2] S[table-format=2.2]}
    \hline
    \multirow{3}{*}{\textbf{Method}} & \multicolumn{4}{c}{\textbf{BEHAVE~\cite{bhatnagar2022behave}}} & \multicolumn{1}{c}{} & \multicolumn{4}{c}{\textbf{InterCap~\cite{huang2022intercap, huang2024intercap}}} \\
    \cline{2-5}
    \cline{7-10}
    & \multicolumn{2}{c}{Amodal Completion} & \multicolumn{2}{c}{Temporal Consistency} & \multicolumn{1}{c}{} & \multicolumn{2}{c}{Amodal Completion} & \multicolumn{2}{c}{Temporal Consistency} \\
    \cline{2-5}
    \cline{7-10}
    \noalign{\vspace{3pt}}
    & {IoU \(\uparrow\)} & {CLIP \(\uparrow\)} & {\parbox{1.2cm}{\centering Warp-err\\(x$10^{-3}$)} \(\downarrow\)} & {TC Score \(\uparrow\)} & {} & {IoU \(\uparrow\)} & {CLIP \(\uparrow\)} & {\parbox{1.2cm}{\centering Warp-err\\(x$10^{-3}$)} \(\downarrow\)} & {TC Score \(\uparrow\)} \\
    \noalign{\vspace{3pt}}
    \hline
    \hline
    Pix2gestalt~\cite{ozguroglu2024pix2gestalt} & {61.29\%} & {26.77} & {141.08} & {95.97} & {} & {64.91\%} & {25.32} & {112.75} & {95.83} \\
    SD Inpainting~\cite{rombach2022high} & {60.81\%} & {27.63} & {9.87} & {96.78} & {} & {45.16\%} & {27.23} & {7.04} & {96.71} \\
    LaMa~\cite{suvorov2022resolution} & {43.54\%} & {26.08} & {6.34} & {96.96} & {} & {56.42\%} & {26.02} & {3.78} & \textbf{98.01} \\
    VDT~\cite{lu2023vdt} & {55.69\%} & {26.48} & \textbf{5.84} & {96.58} & {} & {59.96\%} & {26.11} & {3.42} & {96.39} \\
    \textbf{Ours} (HDM Mask)  & {\textbf{61.75}\%} & \textbf{27.64} & \underline{6.26} & \textbf{97.19} & {} & {\textbf{70.18}\%} & {\textbf{27.65}} & {\textbf{3.22}} & \underline{96.95} \\
    \hline
    \textbf{Ours} (Ground Truth Mask) & {70.90\%} & {28.01} & {6.09} & {98.15} & {} & {74.79\%} & {27.66} & {2.93} & {97.98} \\
    \hline
    \end{tabular}
    % \vspace{10pt}
    % \vspace{0.5em}
    \caption{Quantitative Comparison on BEHAVE~\cite{bhatnagar2022behave} and InterCap~\cite{huang2022intercap, huang2024intercap} for Amodal Completion and Temporal Consistency: Our method achieves consistently strong performance, highlighting its robustness to occlusion and temporal challenges. Bold and underline denote the best and second-best scores.} %Our method outperforms baselines across most metrics, especially under severe occlusions. The “Ground Truth Mask” setting serves as an upper bound.}
    \vspace{-3.0em}
    \label{tab:quantitative}
\end{table*}

\begin{figure}[t]
    \centering
    \includegraphics[width=0.96\linewidth]{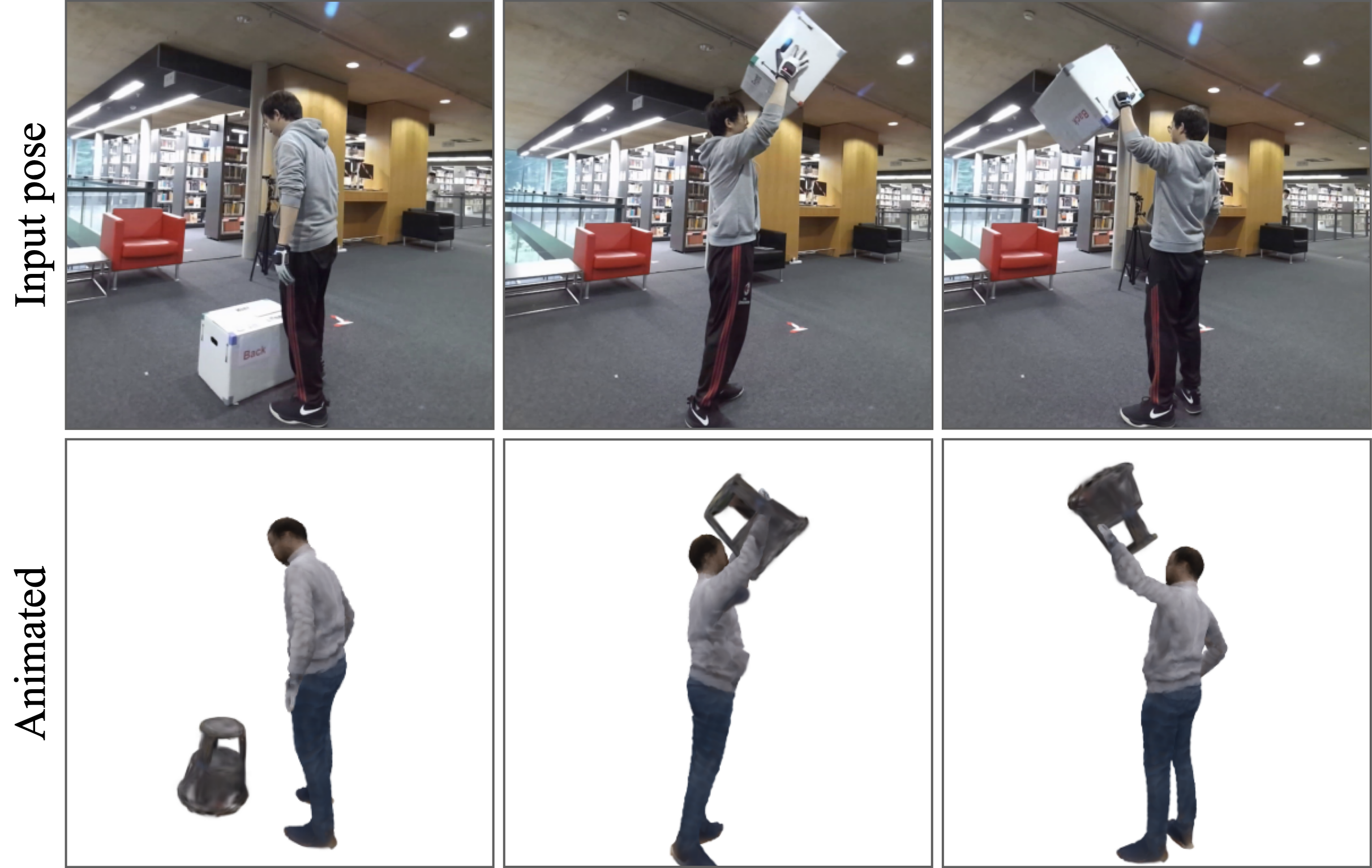}
    \vspace{-1.0em}
    \caption{Animatable 3D Reconstruction of Human–Object Interactions: we train 3DGS~\cite{kerbl3Dgaussians} using temporally consistent amodal completions from our pipeline, sampled at 1 fps (15–47 frames per video). Conditioned on motion trajectories from the input video, our method enables realistic animation of novel human–object pairs while preserving geometry, appearance, and temporal coherence.} % 
    \label{fig:3d}
    \vspace{-2.0em}
\end{figure}

\vspace{-0.7em}
\subsection{Application in 3D Reconstruction} \label{sec:3D_recon}

To demonstrate the utility of our temporally consistent amodal completion in downstream tasks, we reconstruct photo-realistic and animatable 3D human--object interaction scenes from monocular video using 3D Gaussian Splatting (3DGS)~\cite{kerbl3Dgaussians}. Our pipeline produces temporally inpainted frames \(I_{out}^t\), which serve as dense appearance supervision even under severe occlusions. %These completed frames enable high-quality 3D reconstruction from monocular video by revealing full geometry and texture information across time.

%\kwon{What is the difference between $I_{out}^t$ with and without tilde? only difference is that tilde I means croped and recentered. OK} 
For object reconstruction, we follow a GS-Pose~\cite{mei2024gs2pose}, using provided 6-DoF object and camera poses. To ensure consistent scale and visibility, each object in image is cropped and re-centered in every frame $\tilde{I}^t_{out}$. A 3D Gaussian representation \(\mathcal{G}_{obj}\) is then optimized by minimizing a photometric loss between rendered images and inpainted frames:
\vspace{-0.5em}
\begin{equation}
\mathcal{L}_{obj} = \sum_t \mathcal{L}_{photo}\left( R(\mathcal{G}_{obj}, H_t \circ X_t),\; \tilde{I}_{out}^t \right),
\vspace{-0.5em}
\end{equation}
where \(R(\cdot)\) denotes the Gaussian renderer, and \(H_t\), \(X_t\) are the camera and object poses. The \(\mathcal{L}_{photo}\) is a combination of L1 and SSIM terms:
\begin{equation}
\mathcal{L}_{photo}(I_r, I_{gt}) = \| I_r - I_{gt} \|_1 + \lambda \cdot \left(1 - \mathrm{SSIM}(I_r, I_{gt})\right),
\end{equation}
where \(I_r\) is the rendered image, \(I_{\text{gt}}\) is the ground truth image, and \(\lambda\) controls the balance between pixel-wise and perceptual similarity. In our experiments, we set \(\lambda\) = 0.2 to balance these terms effectively.

For human reconstruction, we adopt GaussianAvatar~\cite{hu2024gaussianavatar}, which learns a canonical 3D Gaussian representation and a pose-conditioned deformation field based on SMPL parameters. The model is trained with the same inpainted sequences, allowing robust geometry and appearance recovery even under partial occlusions.

Together, as visualized in~\Cref{fig:3d}, these reconstructions validate that our temporally consistent amodal completion provides a reliable and expressive foundation for photo-realistic, animatable 3D human-object modeling from monocular video.
\vspace{-0.7em}
\section{Experiment}
\label{sec:experiment}

\vspace{-0.3em}
\subsection{Datasets}

\paragraph{\textbf{BEHAVE}} The BEHAVE dataset \cite{bhatnagar2022behave} consists of 321 RGB-D sequences capturing indoor human–object interactions, recorded using 4 Kinect cameras. The test set includes 3 subjects interacts with 20 objects. Among these, we select 18 videos for 18 objects excluding keyboard and basketball due to the lacking of ground truth pose annotation in 30 fps video sequence. For human, we selected 3 videos for 3 subjects in the test set. As a results, we applied our pipeline to approximately 27,000 frames and evaluate these results.

\vspace{-0.5em}
\paragraph{\textbf{InterCap}} InterCap~\cite{huang2022intercap, huang2024intercap} contains 223 RGB-D videos of human-object interactions, captured from 6 distinct viewpoints with 10 subjects and 10 objects. From this dataset, we extract 10 videos that collectively represent all 10 objects.

\vspace{-0.8em}
\subsection{Evaluation Metrics}

\paragraph{\textbf{Amodal Completion}} We evaluate the performance of our amodal completion using two metrics: the CLIP score \cite{radford2021learning} and Intersection over Union (IoU). The CLIP score evaluates the alignment between the generated images and the corresponding category prompts, while IoU measures the overlap between the predicted and groundtruth amodal masks.  Specifically, we calculate the CLIP score for each inpainted frame within tight bounding boxes to minimize the influence of background pixels. For the IoU evaluation, we segment the inpainted frame using SAM2~\cite{ravi2024sam2} to obtain masks and then calculate the overlap with the object's groundtruth masks.

\begin{figure*}[t]
    \centering
    \includegraphics[width=\linewidth]{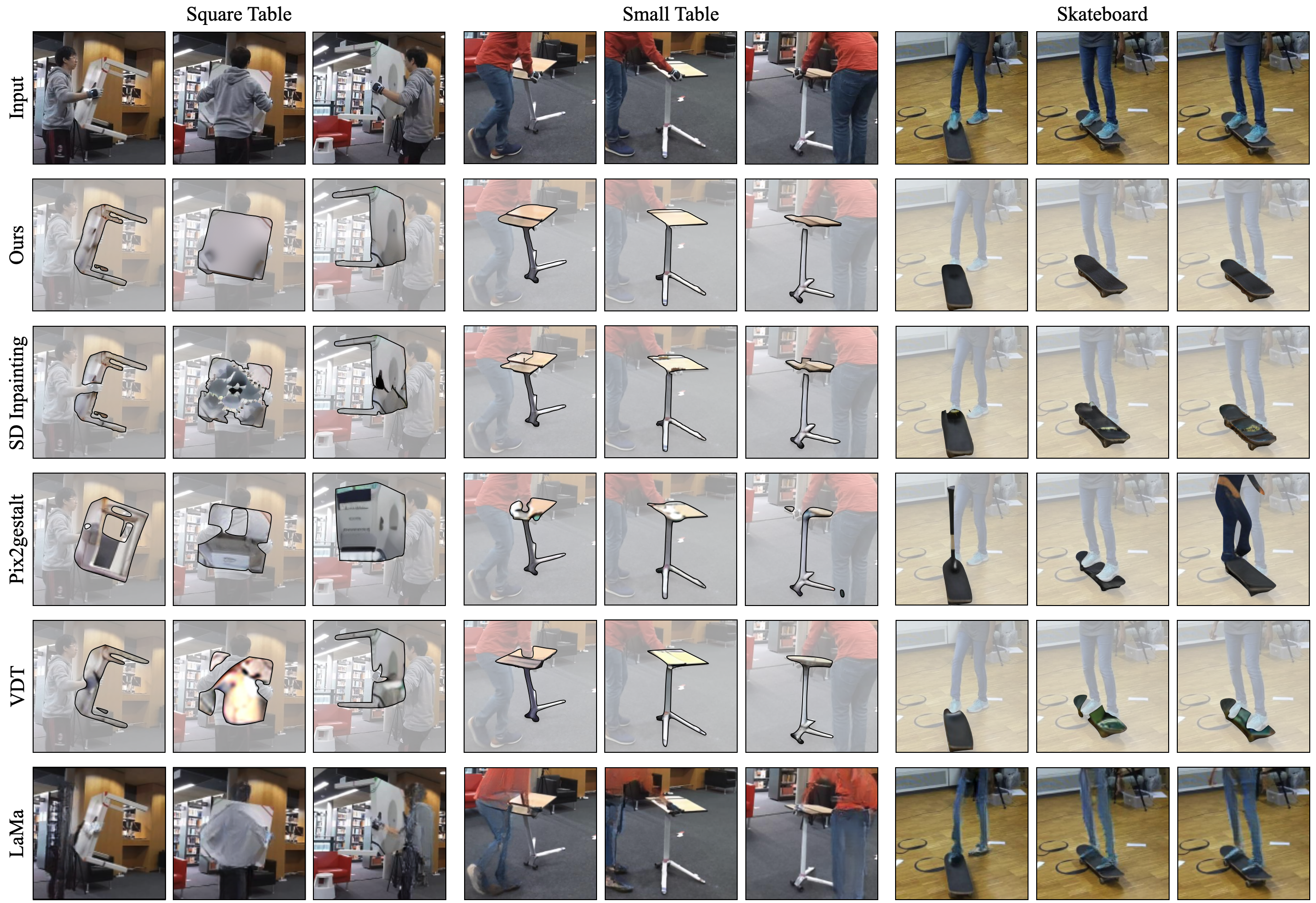}
    \vspace{-2.5em}
    \caption{Qualitative comparison on BEHAVE~\cite{bhatnagar2022behave} (Square Table, Small Table) and InterCap~\cite{huang2022intercap, huang2024intercap} (Skateboard). Our method produces accurate and temporally consistent completions of occluded regions.}
    \label{fig:qualitative}
    \vspace{-1.em}
\end{figure*}

\vspace{-0.5em}
\paragraph{\textbf{Temporal Consistency}}
We assess temporal consistency using two metrics: the Temporal Consistency score (TC score)~\cite{esser2023structure} and the Flow Warping Error~\cite{lai2018learning}. The TC score is computed by extracting CLIP image embeddings for all output video frames and calculating the average cosine similarity between pairs of consecutive frames. Flow Warping Error, commonly used in optical flow estimation and video processing, measures the discrepancy between a warped image and its corresponding target frame. In our approach, we compute optical flows between consecutive ground-truth object masks using SEA-RAFT~\cite{wang2024sea}. We then warp the previously inpainted frame to align with the current frame, and measure the discrepancy between this warped previous frame and the actual current frame. A higher Temporal Consistency Score and a lower Flow Warping Error both indicate better temporal consistency. 

\vspace{-0.7em}
\paragraph{\textbf{3D Reconstruction}} 
Since our method reconstructs only the object or the human—excluding the background—we evaluate reconstruction quality using masked versions of standard metrics: Peak Signal-to-Noise Ratio (PSNR-M), Structural Similarity Index Measure (SSIM-M)~\cite{wang2004image}, and Learned Perceptual Image Patch Similarity (LPIPS-M)~\cite{zhang2018unreasonable}. Inspired by ~\cite{nazarczuk2024aim}, PSNR-M, SSIM-M and LPIPS-M are calculated within tight bounding boxes around the reconstructed regions to minimize the influence of background pixels.

\vspace{-0.8em}
\subsection{Temporally Consistent Amodal Completion} 
\paragraph{\textbf{Quantitative Results}} 
We quantitatively evaluate our approach against several state-of-the-art baselines, including Pix2gestalt~\cite{ozguroglu2024pix2gestalt}, Stable Diffusion Inpainting~\cite{rombach2022high}, LaMa~\cite{suvorov2022resolution}, and VDT~\cite{lu2023vdt}, initialized with random noise. The results are summarized in~\Cref{tab:quantitative}, covering both human and object categories, as the human is treated as one of 19 object classes in our experiments, and only frames with an object occlusion ratio between 15\% and 70\% are considered.
Our method consistently achieves the best performance across amodal completion metrics, IoU and CLIP, on both BEHAVE and InterCap. This indicates a higher fidelity in reconstructing occluded regions, validating the effectiveness of our template-free regioning strategy in complex HOI cases.
To assess ideal condition performance, we report results using ground-truth masks in the last row.

Regarding temporal consistency, our method achieves the lowest warping error among all methods, reflecting superior temporal alignment between consecutive frames. This highlights our model's strength in preserving temporal coherence throughout the video sequence.
Interestingly, although LaMa reports the highest TC score, it ranks lowest in IoU and performs poorly in terms of visual quality. This may be attributed to its tendency to overwrite masked regions with background-matching content, which is especially beneficial in static-camera datasets like BEHAVE and InterCap. In contrast, our method reconstructs occlusions by explicitly leveraging spatial-temporal information from adjacent frames, leading to more semantically faithful and temporally robust completions.
% \vspace{-0.5em}

\vspace{-0.5em}
\paragraph{\textbf{Qualitative Results}} 
\Cref{fig:qualitative} presents a qualitative comparison across various methods on three representative HOI categories: \textit{Square Table}, \textit{Small Table}, and \textit{Skateboard}. Compared to baseline methods, our approach produces semantically faithful and visually coherent reconstructions, especially in the occluded regions.
For \textit{Square Table}, Pix2gestalt generates the implausible shape and VDT fails to complete the proper region. In contrast, our method preserves object integrity with precise geometry and sharper texture completion.
In the \textit{Skateboard} scenario, our method excels in preserving object shape and continuity under occlusion. Competing methods often hallucinate incorrect geometry or texture, especially under relatively simple occlusions caused by human limbs.
Overall, the visual results confirm that our method outperforms prior approaches in generating amodally completed outputs that are both spatially and temporally coherent, reinforcing the strength of our template-free, temporally consistent completion framework.

\vspace{-1.0em}
\subsection{3D Reconstruction}

To the best of our knowledge, this is the first method to enable animatable and photo-realistic 3D reconstruction in human–object interaction (HOI) scenarios through temporally consistent amodal completion. We use the completed monocular video frames generated by our pipeline to train 3D Gaussian Splatting~\cite{kerbl3Dgaussians} (3DGS), enabling high-quality reconstruction from monocular inputs. For training, we extract 40 to 47 frames per sequence at 1 fps, a sparse set of inpainted frames.

\begin{figure}[t]
    \centering
    \includegraphics[width=\linewidth]{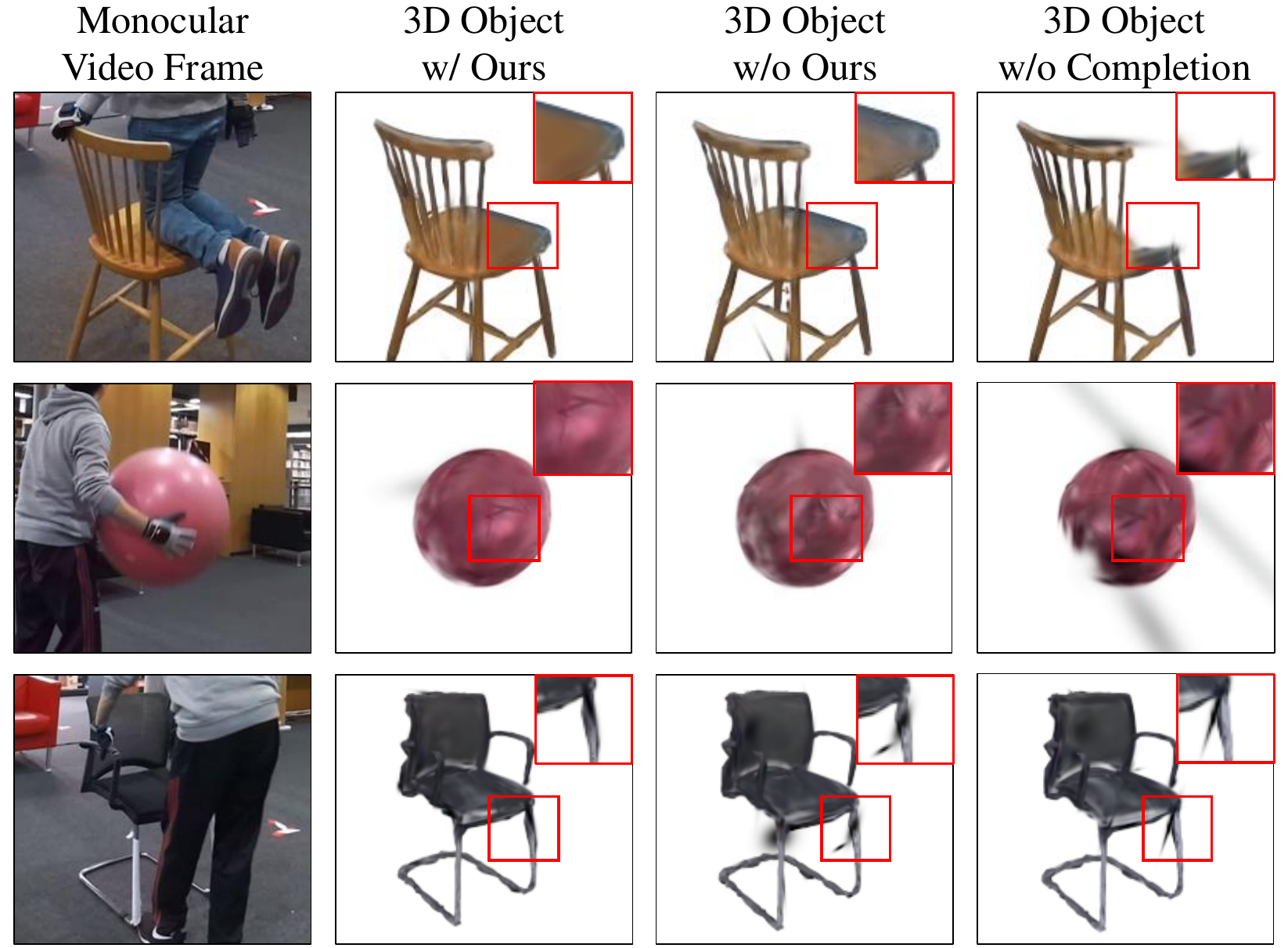}
    \vspace{-2.5em}
    \caption{Qualitative results of 3D Reconstruction: Our method (left) produces clearer geometry and fewer artifacts than without our method (middle) or no completion (right).}
    \vspace{-1.0em}
    \label{fig:3d2}
\end{figure}

We compare three settings: (1) our full pipeline with temporally-aware amodal completion, (2) Stable Diffusion inpainting without our temporal modules, and (3) the original input sequence without any completion. As shown in~\Cref{tab:3d_metric}, our method achieves the highest PSNR-M, SSIM-M and LPIPS-M, outperforming all comparison methods.
As illustrated in~\Cref{fig:3d2}, our approach improves both geometric accuracy and temporal consistency, which are crucial for smooth and realistic novel-view rendering. Furthermore, when conditioned on SMPL body poses or 6D object trajectories, the trained model produces photo-realistic animations of both humans and objects, highlighting the robustness and completeness of our pipeline.
In contrast, without our temporally consistent amodal completion, occluded regions remain unresolved and frame-to-frame inconsistencies persist, leading to degraded 3D reconstruction quality.

\begin{table}
  \centering
  \begin{tabular}{c c c c}
    \toprule
    \textbf{Method} & PSNR-M \(\uparrow\) & SSIM-M \(\uparrow\) & LPIPS-M \(\downarrow\)\\ 
    \hline\hline
    w/o Completion   & {13.48}    & {0.5679}  & {0.3604} \\
    SD Inpainting~\cite{rombach2022high}   & 15.38    & 0.6186  & 0.2455 \\
    \textbf{Ours}   & \textbf{15.43}    & \textbf{0.6240} & \textbf{0.2383} \\
    \hline
  \end{tabular}
  \vspace{1pt}
    \caption{Quantitative results on 3D reconstruction from the BEHAVE dataset~\cite{bhatnagar2022behave}: We report the metrics, computed within tight bounding boxes around humans or objects.}%,Our method outperforms all baselines across metrics. demonstrating accurate and temporally consistent amodal completion.}
  \label{tab:3d_metric}
  \vspace{-2.2em}
\end{table}

\begin{table}
  \centering
  \renewcommand{\arraystretch}{1.3}
  \begin{tabular}{c c c c c c}
    \toprule
    \textbf{\parbox{1.2cm}{\centering Feature\\Warp.}} & \textbf{\parbox{1.2cm}{\centering Cross\\Attn.}} & {IoU \(\uparrow\)} & {CLIP \(\uparrow\)} & {\parbox{1.2cm}{\centering Warp-err\\(x$10^{-3}$)} \(\downarrow\)} & {TC \(\uparrow\)} \\
    % \textbf{Feature Warping} & \textbf{Cross Attention} & {IoU} & {CLIP} & {Warp-err (x$10^{-3}$)} & {TC Score} \\
    \hline\hline
    \ding{55} & \ding{55} & {60.81} & {27.63} & {9.87} & {96.78} \\
    \ding{55} & \ding{51} & {60.82} & {27.64} & {6.56} & {97.06} \\
    \ding{51} & \ding{55} & {58.35} & {27.60} & \textbf{6.23} & {97.16} \\
    \ding{51} & \ding{51} & \textbf{61.75} & \textbf{27.64} & \underline{6.26} & \textbf{97.19} \\
    \hline
  \end{tabular}
  \vspace{1pt}
  \caption{Ablation Studies on temporal consistency strategy for on BEHAVE dataset~\cite{bhatnagar2022behave}. Bold
  and underline denote the best and second-best scores.}
  \label{tab:ablation}
  \vspace{-3.0em}
\end{table}

\vspace{-0.5em}
\subsection{Ablation Study}
To assess the contributions of each component in our temporal consistency framework, we conduct ablation studies on the BEHAVE dataset, as shown in~\Cref{tab:ablation}. We analyze the impact of two key components: feature warping and cross-frame attention. Without either feature warping or cross attention (first row), the model performs reasonably well, but shows higher warping error and lower temporal consistency. Introducing cross attention alone (second row) slightly improves all metrics, particularly reducing the warping error. Interestingly, feature warping alone (third row) leads to the lowest warping error (6.23) and improved TC score, demonstrating its effectiveness in aligning features across time. However, it degrades amodal completion accuracy (IoU).
Combining both feature warping and cross attention (last row) yields the best overall performance, achieving the highest IoU (61.75), CLIP score (27.64), and TC score (97.19). This confirms that both components are complementary and essential for generating temporally consistent and semantically meaningful completions.

We conduct an additional ablation study to evaluate the effectiveness of our template-free occlusion identification strategy, as shown in~\Cref{tab:ablation}. Specifically, we compare our method against a baseline that relies on predefined human masks to guide the inpainting process. Our approach derives the occlusion mask $M_{\text{occlusion}}$ by combining 2D segmentation with 3D point cloud projection, leading to significantly better performance in terms of IoU (61.75 vs. 53.77) and warping error (6.26 vs. 8.90). These results highlight our method’s superior ability to accurately localize occluded regions and maintain temporal coherence. Although the human-mask baseline yields a slightly higher CLIP score and TC Score, this is likely due to its tendency to inpaint overly large regions, which degrades spatial accuracy and temporal consistency, as reflected in the lower IoU and higher warping error. Qualitative results of the ablation study are provided in the supplementary material.

\vspace{-0.5em}
\section{Discussion and Limitations}
\label{sec:discussion_limitation}
\begin{table}
  \centering
  \renewcommand{\arraystretch}{1.3}
  \begin{tabular}{c c c c c}
    \toprule
    \textbf{Mask} & {IoU \(\uparrow\)} & {CLIP \(\uparrow\)} & {\parbox{1.2cm}{\centering Warp-err\\(x$10^{-3}$)} \(\downarrow\)} & {TC \(\uparrow\)} \\
    % \textbf{Feature Warping} & \textbf{Cross Attention} & {IoU} & {CLIP} & {Warp-err (x$10^{-3}$)} & {TC Score} \\
    \hline\hline
    {Human Mask} & {53.77} & \textbf{28.31} & {8.90} & \textbf{97.60} \\
    \textbf{Ours} (\(M_{occlusion}\)) & \textbf{61.75} & {27.64} & \textbf{6.26} & {97.19} \\
    \hline
  \end{tabular}
  \vspace{1pt}
  \caption{Ablation Studies on Template-free Occlusion Identification strategy on BEHAVE dataset~\cite{bhatnagar2022behave}.}
  \label{tab:mask_ablation}
  \vspace{-3.0em}
\end{table}

Dynamic human–object interactions present significant challenges for optical flow-based shape estimation, particularly due to non-linear motion and frequent occlusions. Our approach leverages HDM for amodal completion, enabling the reconstruction of full object shapes—including occluded regions. However, the method may not always accurately infer the geometry of unseen parts. Future directions include integrating multi-view geometry techniques, such as Structure-from-Motion (SfM) and Multi-View Stereo (MVS), to densify and refine reconstructions, thereby enhancing the robustness of our pipeline.

Currently, our method assumes that each input video contains only a single human and a single object. While effective for controlled scenarios, this assumption limits applicability in more complex settings, such as multi-person interactions or interactions involving multiple objects. Extending the framework to handle such cases is a promising direction for future work.

Additionally, our method relies heavily on the inpainting capabilities of the Stable Diffusion model, which may struggle to accurately reconstruct objects or appearances that were not well represented during its training. We also observed that the quality of inpainting is highly sensitive to the accuracy of the occlusion masks. In particular, when the masks fail to fully exclude occluding regions—leaving behind small residual artifacts—the inpainting output is noticeably degraded. These limitations highlight the critical importance of precise occlusion localization, especially in scenes involving fine-grained interactions or partial visibility.
% \vspace{-0.5em}
\vspace{-0.5em}
\section{Conclusion}\label{sec:conclusion}
We proposed a novel framework for reconstructing dynamic human–object interactions from monocular video, with a focus on addressing occlusions and temporal inconsistency. Our method achieves state-of-the-art performance across multiple benchmarks, demonstrating both quantitative superiority and qualitative robustness. It also generalizes well to real-world scenarios. Notably, by integrating our approach with 3D Gaussian Splatting, we enable the generation of photorealistic and animatable 3D reconstructions from monocular input, supporting advanced applications such as novel-view and novel-pose synthesis in complex HOI scenes.

\clearpage

\begin{acks}
We wish to thank all the reviewers for their invaluable feedback.
This work is partially supported by the NSF under the Future of Work at the
Human-Technology Frontier (FW-HTF) 1839971 and Partnership
for Innovation: Technology Transfer (PFI-TT) 2329804. Additional support for this work is provided by the Culture, Sports, and Tourism R\&D Program through the Korea Creative Content Agency grant funded by the Ministry of Culture, Sports and Tourism in 2024~(International Collaborative Research and Global Talent Development for the Development of Copyright Management and Protection Technologies for Generative AI, RS-2024-00345025), Institute of Information \& communications Technology Planning \& Evaluation (IITP) grant funded by the Korea government (MSIT) (No. RS-2019-II190079, Artificial Intelligence Graduate School Program (Korea University).
We also acknowledge the Feddersen Distinguished Professorship Funds and
a gift from Thomas J. Malott. Any opinions, findings, and
conclusions expressed in this material are those of the authors and
do not necessarily reflect the views of the funding agency.
\end{acks}

\bibliographystyle{ACM-Reference-Format}
\balance
\bibliography{main}


\begin{thebibliography}{56}

%%% the Web address.

\ifx \showCODEN    \undefined \def \showCODEN     #1{\unskip}     \fi
\ifx \showISBNx    \undefined \def \showISBNx     #1{\unskip}     \fi
\ifx \showISBNxiii \undefined \def \showISBNxiii  #1{\unskip}     \fi
\ifx \showISSN     \undefined \def \showISSN      #1{\unskip}     \fi
\ifx \showLCCN     \undefined \def \showLCCN      #1{\unskip}     \fi
\ifx \shownote     \undefined \def \shownote      #1{#1}          \fi
\ifx \showarticletitle \undefined \def \showarticletitle #1{#1}   \fi
\ifx \showURL      \undefined \def \showURL       {\relax}        \fi
% invisible to TeX
\providecommand\bibfield[2]{#2}
\providecommand\bibinfo[2]{#2}
\providecommand\natexlab[1]{#1}
\providecommand\showeprint[2][]{arXiv:#2}

\bibitem[Bellock({[n.\,d.]})]%
        {bellockk_alphashape}
\bibfield{author}{\bibinfo{person}{K. Bellock}.} \bibinfo{year}{[n.\,d.]}\natexlab{}.
\newblock \bibinfo{title}{alphashape}.
\newblock \bibinfo{howpublished}{\url{https://github.com/bellockk/alphashape}}.
\newblock
\newblock
\shownote{GitHub repository, accessed 2025-04-11}.


\bibitem[Bhatnagar et~al\mbox{.}(2022)]%
        {bhatnagar2022behave}
\bibfield{author}{\bibinfo{person}{Bharat~Lal Bhatnagar}, \bibinfo{person}{Xianghui Xie}, \bibinfo{person}{Ilya~A Petrov}, \bibinfo{person}{Cristian Sminchisescu}, \bibinfo{person}{Christian Theobalt}, {and} \bibinfo{person}{Gerard Pons-Moll}.} \bibinfo{year}{2022}\natexlab{}.
\newblock \showarticletitle{Behave: Dataset and method for tracking human object interactions}. In \bibinfo{booktitle}{\emph{Proceedings of the IEEE/CVF Conference on Computer Vision and Pattern Recognition}}. \bibinfo{pages}{15935--15946}.
\newblock


\bibitem[Chen et~al\mbox{.}(2023)]%
        {chen2023amodal3d}
\bibfield{author}{\bibinfo{person}{A. Chen}, \bibinfo{person}{B. Smith}, {and} \bibinfo{person}{C. Lee}.} \bibinfo{year}{2023}\natexlab{}.
\newblock \showarticletitle{Amodal 3D Shape from Partial Views}. In \bibinfo{booktitle}{\emph{Proceedings of the IEEE/CVF International Conference on Computer Vision (ICCV)}}. \bibinfo{pages}{4567--4576}.
\newblock


\bibitem[Chi et~al\mbox{.}(2024)]%
        {chi2024m2d2m}
\bibfield{author}{\bibinfo{person}{Seunggeun Chi}, \bibinfo{person}{Hyung-gun Chi}, \bibinfo{person}{Hengbo Ma}, \bibinfo{person}{Nakul Agarwal}, \bibinfo{person}{Faizan Siddiqui}, \bibinfo{person}{Karthik Ramani}, {and} \bibinfo{person}{Kwonjoon Lee}.} \bibinfo{year}{2024}\natexlab{}.
\newblock \showarticletitle{M2d2m: Multi-motion generation from text with discrete diffusion models}. In \bibinfo{booktitle}{\emph{European conference on computer vision}}. Springer, \bibinfo{pages}{18--36}.
\newblock


\bibitem[Chi et~al\mbox{.}(2025)]%
        {chi2025contactawareamodalcompletionhumanobject}
\bibfield{author}{\bibinfo{person}{Seunggeun Chi}, \bibinfo{person}{Enna Sachdeva}, \bibinfo{person}{Pin-Hao Huang}, {and} \bibinfo{person}{Kwonjoon Lee}.} \bibinfo{year}{2025}\natexlab{}.
\newblock \bibinfo{title}{Contact-Aware Amodal Completion for Human-Object Interaction via Multi-Regional Inpainting}.
\newblock
\showeprint[arxiv]{2508.00427}~[cs.CV]
\urldef\tempurl%
\url{https://arxiv.org/abs/2508.00427}
\showURL{%
\tempurl}


\bibitem[Cong et~al\mbox{.}(2023)]%
        {cong2023flatten}
\bibfield{author}{\bibinfo{person}{Yuren Cong}, \bibinfo{person}{Mengmeng Xu}, \bibinfo{person}{Christian Simon}, \bibinfo{person}{Shoufa Chen}, \bibinfo{person}{Jiawei Ren}, \bibinfo{person}{Yanping Xie}, \bibinfo{person}{Juan-Manuel Perez-Rua}, \bibinfo{person}{Bodo Rosenhahn}, \bibinfo{person}{Tao Xiang}, {and} \bibinfo{person}{Sen He}.} \bibinfo{year}{2023}\natexlab{}.
\newblock \showarticletitle{Flatten: optical flow-guided attention for consistent text-to-video editing}.
\newblock \bibinfo{journal}{\emph{arXiv preprint arXiv:2310.05922}} (\bibinfo{year}{2023}).
\newblock


\bibitem[Esser et~al\mbox{.}(2023)]%
        {esser2023structure}
\bibfield{author}{\bibinfo{person}{Patrick Esser}, \bibinfo{person}{Johnathan Chiu}, \bibinfo{person}{Parmida Atighehchian}, \bibinfo{person}{Jonathan Granskog}, {and} \bibinfo{person}{Anastasis Germanidis}.} \bibinfo{year}{2023}\natexlab{}.
\newblock \showarticletitle{Structure and content-guided video synthesis with diffusion models}. In \bibinfo{booktitle}{\emph{Proceedings of the IEEE/CVF international conference on computer vision}}. \bibinfo{pages}{7346--7356}.
\newblock


\bibitem[Fu et~al\mbox{.}(2024)]%
        {Fu_2024_CVPR}
\bibfield{author}{\bibinfo{person}{Yang Fu}, \bibinfo{person}{Sifei Liu}, \bibinfo{person}{Amey Kulkarni}, \bibinfo{person}{Jan Kautz}, \bibinfo{person}{Alexei~A. Efros}, {and} \bibinfo{person}{Xiaolong Wang}.} \bibinfo{year}{2024}\natexlab{}.
\newblock \showarticletitle{COLMAP-Free 3D Gaussian Splatting}. In \bibinfo{booktitle}{\emph{Proceedings of the IEEE/CVF Conference on Computer Vision and Pattern Recognition (CVPR)}}. \bibinfo{pages}{20796--20805}.
\newblock


\bibitem[Geyer et~al\mbox{.}(2023)]%
        {geyer2023tokenflow}
\bibfield{author}{\bibinfo{person}{Michal Geyer}, \bibinfo{person}{Omer Bar-Tal}, \bibinfo{person}{Shai Bagon}, {and} \bibinfo{person}{Tali Dekel}.} \bibinfo{year}{2023}\natexlab{}.
\newblock \showarticletitle{Tokenflow: Consistent diffusion features for consistent video editing}.
\newblock \bibinfo{journal}{\emph{arXiv preprint arXiv:2307.10373}} (\bibinfo{year}{2023}).
\newblock


\bibitem[Hu et~al\mbox{.}(2024)]%
        {hu2024gaussianavatar}
\bibfield{author}{\bibinfo{person}{Liangxiao Hu}, \bibinfo{person}{Hongwen Zhang}, \bibinfo{person}{Yuxiang Zhang}, \bibinfo{person}{Boyao Zhou}, \bibinfo{person}{Boning Liu}, \bibinfo{person}{Shengping Zhang}, {and} \bibinfo{person}{Liqiang Nie}.} \bibinfo{year}{2024}\natexlab{}.
\newblock \showarticletitle{Gaussianavatar: Towards realistic human avatar modeling from a single video via animatable 3d gaussians}. In \bibinfo{booktitle}{\emph{Proceedings of the IEEE/CVF conference on computer vision and pattern recognition}}. \bibinfo{pages}{634--644}.
\newblock


\bibitem[Huang et~al\mbox{.}(2022)]%
        {huang2022intercap}
\bibfield{author}{\bibinfo{person}{Yinghao Huang}, \bibinfo{person}{Omid Taheri}, \bibinfo{person}{Michael~J Black}, {and} \bibinfo{person}{Dimitrios Tzionas}.} \bibinfo{year}{2022}\natexlab{}.
\newblock \showarticletitle{InterCap: Joint markerless 3D tracking of humans and objects in interaction}. In \bibinfo{booktitle}{\emph{DAGM German Conference on Pattern Recognition}}. Springer, \bibinfo{pages}{281--299}.
\newblock


\bibitem[Huang et~al\mbox{.}(2024)]%
        {huang2024intercap}
\bibfield{author}{\bibinfo{person}{Yinghao Huang}, \bibinfo{person}{Omid Taheri}, \bibinfo{person}{Michael~J Black}, {and} \bibinfo{person}{Dimitrios Tzionas}.} \bibinfo{year}{2024}\natexlab{}.
\newblock \showarticletitle{InterCap: Joint Markerless 3D Tracking of Humans and Objects in Interaction from Multi-view RGB-D Images}.
\newblock \bibinfo{journal}{\emph{International Journal of Computer Vision}} (\bibinfo{year}{2024}), \bibinfo{pages}{1--16}.
\newblock


\bibitem[Jain et~al\mbox{.}(2021)]%
        {jain2021putting}
\bibfield{author}{\bibinfo{person}{Ajay Jain}, \bibinfo{person}{Matthew Tancik}, {and} \bibinfo{person}{Pieter Abbeel}.} \bibinfo{year}{2021}\natexlab{}.
\newblock \showarticletitle{Putting nerf on a diet: Semantically consistent few-shot view synthesis}. In \bibinfo{booktitle}{\emph{Proceedings of the IEEE/CVF International Conference on Computer Vision}}. \bibinfo{pages}{5885--5894}.
\newblock


\bibitem[Jeong et~al\mbox{.}(2022)]%
        {jeong2022imposing}
\bibfield{author}{\bibinfo{person}{Jisoo Jeong}, \bibinfo{person}{Jamie~Menjay Lin}, \bibinfo{person}{Fatih Porikli}, {and} \bibinfo{person}{Nojun Kwak}.} \bibinfo{year}{2022}\natexlab{}.
\newblock \showarticletitle{Imposing consistency for optical flow estimation}. In \bibinfo{booktitle}{\emph{Proceedings of the IEEE/CVF conference on Computer Vision and Pattern Recognition}}. \bibinfo{pages}{3181--3191}.
\newblock


\bibitem[Jiang et~al\mbox{.}(2022)]%
        {jiang2022neuman}
\bibfield{author}{\bibinfo{person}{Wei Jiang}, \bibinfo{person}{Kwang~Moo Yi}, \bibinfo{person}{Golnoosh Samei}, \bibinfo{person}{Oncel Tuzel}, {and} \bibinfo{person}{Anurag Ranjan}.} \bibinfo{year}{2022}\natexlab{}.
\newblock \showarticletitle{Neuman: Neural human radiance field from a single video}. In \bibinfo{booktitle}{\emph{European Conference on Computer Vision}}. Springer, \bibinfo{pages}{402--418}.
\newblock


\bibitem[Kerbl et~al\mbox{.}(2023)]%
        {kerbl3Dgaussians}
\bibfield{author}{\bibinfo{person}{Bernhard Kerbl}, \bibinfo{person}{Georgios Kopanas}, \bibinfo{person}{Thomas Leimk{\"u}hler}, {and} \bibinfo{person}{George Drettakis}.} \bibinfo{year}{2023}\natexlab{}.
\newblock \showarticletitle{3D Gaussian Splatting for Real-Time Radiance Field Rendering}.
\newblock \bibinfo{journal}{\emph{ACM Transactions on Graphics}} \bibinfo{volume}{42}, \bibinfo{number}{4} (\bibinfo{date}{July} \bibinfo{year}{2023}).
\newblock
\urldef\tempurl%
\url{https://repo-sam.inria.fr/fungraph/3d-gaussian-splatting/}
\showURL{%
\tempurl}


\bibitem[Kim et~al\mbox{.}(2023)]%
        {kim2023monocular}
\bibfield{author}{\bibinfo{person}{M. Kim}, \bibinfo{person}{J. Park}, {and} \bibinfo{person}{K. Lee}.} \bibinfo{year}{2023}\natexlab{}.
\newblock \showarticletitle{Monocular Differentiable Rendering for Self-Supervised 3D Amodal Masks}. In \bibinfo{booktitle}{\emph{Proceedings of the IEEE/CVF International Conference on Computer Vision (ICCV)}}. \bibinfo{pages}{789--798}.
\newblock


\bibitem[Kingma et~al\mbox{.}(2013)]%
        {kingma2013auto}
\bibfield{author}{\bibinfo{person}{Diederik~P Kingma}, \bibinfo{person}{Max Welling}, {et~al\mbox{.}}} \bibinfo{year}{2013}\natexlab{}.
\newblock \bibinfo{title}{Auto-encoding variational bayes}.
\newblock


\bibitem[Kocabas et~al\mbox{.}(2024)]%
        {kocabas2024hugs}
\bibfield{author}{\bibinfo{person}{Muhammed Kocabas}, \bibinfo{person}{Jen-Hao~Rick Chang}, \bibinfo{person}{James Gabriel}, \bibinfo{person}{Oncel Tuzel}, {and} \bibinfo{person}{Anurag Ranjan}.} \bibinfo{year}{2024}\natexlab{}.
\newblock \showarticletitle{Hugs: Human gaussian splats}. In \bibinfo{booktitle}{\emph{Proceedings of the IEEE/CVF conference on computer vision and pattern recognition}}. \bibinfo{pages}{505--515}.
\newblock


\bibitem[Lai et~al\mbox{.}(2018)]%
        {lai2018learning}
\bibfield{author}{\bibinfo{person}{Wei-Sheng Lai}, \bibinfo{person}{Jia-Bin Huang}, \bibinfo{person}{Oliver Wang}, \bibinfo{person}{Eli Shechtman}, \bibinfo{person}{Ersin Yumer}, {and} \bibinfo{person}{Ming-Hsuan Yang}.} \bibinfo{year}{2018}\natexlab{}.
\newblock \showarticletitle{Learning blind video temporal consistency}. In \bibinfo{booktitle}{\emph{Proceedings of the European conference on computer vision (ECCV)}}. \bibinfo{pages}{170--185}.
\newblock


\bibitem[Lee et~al\mbox{.}(2025)]%
        {lee2025editsplat}
\bibfield{author}{\bibinfo{person}{Dong~In Lee}, \bibinfo{person}{Hyeongcheol Park}, \bibinfo{person}{Jiyoung Seo}, \bibinfo{person}{Eunbyung Park}, \bibinfo{person}{Hyunje Park}, \bibinfo{person}{Ha~Dam Baek}, \bibinfo{person}{Sangheon Shin}, \bibinfo{person}{Sangmin Kim}, {and} \bibinfo{person}{Sangpil Kim}.} \bibinfo{year}{2025}\natexlab{}.
\newblock \showarticletitle{Editsplat: Multi-view fusion and attention-guided optimization for view-consistent 3d scene editing with 3d gaussian splatting}. In \bibinfo{booktitle}{\emph{Proceedings of the Computer Vision and Pattern Recognition Conference}}. \bibinfo{pages}{11135--11145}.
\newblock


\bibitem[Lee et~al\mbox{.}(2024)]%
        {lee2024guess}
\bibfield{author}{\bibinfo{person}{Inhee Lee}, \bibinfo{person}{Byungjun Kim}, {and} \bibinfo{person}{Hanbyul Joo}.} \bibinfo{year}{2024}\natexlab{}.
\newblock \showarticletitle{Guess the unseen: Dynamic 3d scene reconstruction from partial 2d glimpses}. In \bibinfo{booktitle}{\emph{Proceedings of the IEEE/CVF conference on computer vision and pattern recognition}}. \bibinfo{pages}{1062--1071}.
\newblock


\bibitem[Li et~al\mbox{.}(2022)]%
        {li2022compositional}
\bibfield{author}{\bibinfo{person}{P. Li}, \bibinfo{person}{Q. Zhang}, {and} \bibinfo{person}{R. Others}.} \bibinfo{year}{2022}\natexlab{}.
\newblock \showarticletitle{Compositional Models for Amodal Layout Completion}. In \bibinfo{booktitle}{\emph{Proceedings of the IEEE/CVF Conference on Computer Vision and Pattern Recognition (CVPR)}}. \bibinfo{pages}{2345--2354}.
\newblock


\bibitem[Lin et~al\mbox{.}(2024)]%
        {lin2024vastgaussian}
\bibfield{author}{\bibinfo{person}{Jiaqi Lin}, \bibinfo{person}{Zhihao Li}, \bibinfo{person}{Xiao Tang}, \bibinfo{person}{Jianzhuang Liu}, \bibinfo{person}{Shiyong Liu}, \bibinfo{person}{Jiayue Liu}, \bibinfo{person}{Yangdi Lu}, \bibinfo{person}{Xiaofei Wu}, \bibinfo{person}{Songcen Xu}, \bibinfo{person}{Youliang Yan}, {et~al\mbox{.}}} \bibinfo{year}{2024}\natexlab{}.
\newblock \showarticletitle{Vastgaussian: Vast 3d gaussians for large scene reconstruction}. In \bibinfo{booktitle}{\emph{Proceedings of the IEEE/CVF Conference on Computer Vision and Pattern Recognition}}. \bibinfo{pages}{5166--5175}.
\newblock


\bibitem[Lu et~al\mbox{.}(2023)]%
        {lu2023vdt}
\bibfield{author}{\bibinfo{person}{Haoyu Lu}, \bibinfo{person}{Guoxing Yang}, \bibinfo{person}{Nanyi Fei}, \bibinfo{person}{Yuqi Huo}, \bibinfo{person}{Zhiwu Lu}, \bibinfo{person}{Ping Luo}, {and} \bibinfo{person}{Mingyu Ding}.} \bibinfo{year}{2023}\natexlab{}.
\newblock \showarticletitle{Vdt: General-purpose video diffusion transformers via mask modeling}.
\newblock \bibinfo{journal}{\emph{arXiv preprint arXiv:2305.13311}} (\bibinfo{year}{2023}).
\newblock


\bibitem[Lu et~al\mbox{.}(2024)]%
        {scaffoldgs}
\bibfield{author}{\bibinfo{person}{Tao Lu}, \bibinfo{person}{Mulin Yu}, \bibinfo{person}{Linning Xu}, \bibinfo{person}{Yuanbo Xiangli}, \bibinfo{person}{Limin Wang}, \bibinfo{person}{Dahua Lin}, {and} \bibinfo{person}{Bo Dai}.} \bibinfo{year}{2024}\natexlab{}.
\newblock \showarticletitle{Scaffold-gs: Structured 3d gaussians for view-adaptive rendering}. In \bibinfo{booktitle}{\emph{Proceedings of the IEEE/CVF Conference on Computer Vision and Pattern Recognition}}. \bibinfo{pages}{20654--20664}.
\newblock


\bibitem[Mei et~al\mbox{.}(2024)]%
        {mei2024gs2pose}
\bibfield{author}{\bibinfo{person}{Jilan Mei}, \bibinfo{person}{Junbo Li}, {and} \bibinfo{person}{Cai Meng}.} \bibinfo{year}{2024}\natexlab{}.
\newblock \showarticletitle{GS2Pose: Tow-stage 6D Object Pose Estimation Guided by Gaussian Splatting}.
\newblock \bibinfo{journal}{\emph{arXiv preprint arXiv:2411.03807}} (\bibinfo{year}{2024}).
\newblock


\bibitem[Mildenhall et~al\mbox{.}(2021)]%
        {mildenhall2021nerf}
\bibfield{author}{\bibinfo{person}{Ben Mildenhall}, \bibinfo{person}{Pratul~P Srinivasan}, \bibinfo{person}{Matthew Tancik}, \bibinfo{person}{Jonathan~T Barron}, \bibinfo{person}{Ravi Ramamoorthi}, {and} \bibinfo{person}{Ren Ng}.} \bibinfo{year}{2021}\natexlab{}.
\newblock \showarticletitle{Nerf: Representing scenes as neural radiance fields for view synthesis}.
\newblock \bibinfo{journal}{\emph{Commun. ACM}} \bibinfo{volume}{65}, \bibinfo{number}{1} (\bibinfo{year}{2021}), \bibinfo{pages}{99--106}.
\newblock


\bibitem[Nazarczuk et~al\mbox{.}(2024)]%
        {nazarczuk2024aim}
\bibfield{author}{\bibinfo{person}{Michal Nazarczuk}, \bibinfo{person}{Thomas Tanay}, \bibinfo{person}{Sibi Catley-Chandar}, \bibinfo{person}{Richard Shaw}, \bibinfo{person}{Radu Timofte}, {and} \bibinfo{person}{Eduardo P{\'e}rez-Pellitero}.} \bibinfo{year}{2024}\natexlab{}.
\newblock \showarticletitle{AIM 2024 sparse neural rendering challenge: Dataset and benchmark}.
\newblock \bibinfo{journal}{\emph{arXiv preprint arXiv:2409.15041}} (\bibinfo{year}{2024}).
\newblock


\bibitem[Nguyen et~al\mbox{.}(2022)]%
        {nguyen2022disentangled}
\bibfield{author}{\bibinfo{person}{H. Nguyen}, \bibinfo{person}{T. Davis}, {and} \bibinfo{person}{X. Xu}.} \bibinfo{year}{2022}\natexlab{}.
\newblock \showarticletitle{Learning Disentangled Shape-Texture for Amodal Completion}. In \bibinfo{booktitle}{\emph{Advances in Neural Information Processing Systems (NeurIPS)}}. \bibinfo{pages}{1--12}.
\newblock


\bibitem[Ozguroglu et~al\mbox{.}(2024)]%
        {ozguroglu2024pix2gestalt}
\bibfield{author}{\bibinfo{person}{Ege Ozguroglu}, \bibinfo{person}{Ruoshi Liu}, \bibinfo{person}{D{\'\i}dac Sur{\'\i}s}, \bibinfo{person}{Dian Chen}, \bibinfo{person}{Achal Dave}, \bibinfo{person}{Pavel Tokmakov}, {and} \bibinfo{person}{Carl Vondrick}.} \bibinfo{year}{2024}\natexlab{}.
\newblock \showarticletitle{pix2gestalt: Amodal segmentation by synthesizing wholes}. In \bibinfo{booktitle}{\emph{2024 IEEE/CVF Conference on Computer Vision and Pattern Recognition (CVPR)}}. IEEE Computer Society, \bibinfo{pages}{3931--3940}.
\newblock


\bibitem[Poole et~al\mbox{.}(2022)]%
        {poole2022dreamfusion}
\bibfield{author}{\bibinfo{person}{Ben Poole}, \bibinfo{person}{Ajay Jain}, \bibinfo{person}{Jonathan~T Barron}, {and} \bibinfo{person}{Ben Mildenhall}.} \bibinfo{year}{2022}\natexlab{}.
\newblock \showarticletitle{Dreamfusion: Text-to-3d using 2d diffusion}.
\newblock \bibinfo{journal}{\emph{arXiv preprint arXiv:2209.14988}} (\bibinfo{year}{2022}).
\newblock


\bibitem[Radford et~al\mbox{.}(2021)]%
        {radford2021learning}
\bibfield{author}{\bibinfo{person}{Alec Radford}, \bibinfo{person}{Jong~Wook Kim}, \bibinfo{person}{Chris Hallacy}, \bibinfo{person}{Aditya Ramesh}, \bibinfo{person}{Gabriel Goh}, \bibinfo{person}{Sandhini Agarwal}, \bibinfo{person}{Girish Sastry}, \bibinfo{person}{Amanda Askell}, \bibinfo{person}{Pamela Mishkin}, \bibinfo{person}{Jack Clark}, {et~al\mbox{.}}} \bibinfo{year}{2021}\natexlab{}.
\newblock \showarticletitle{Learning transferable visual models from natural language supervision}. In \bibinfo{booktitle}{\emph{International conference on machine learning}}. PmLR, \bibinfo{pages}{8748--8763}.
\newblock


\bibitem[Ravi et~al\mbox{.}(2024)]%
        {ravi2024sam2}
\bibfield{author}{\bibinfo{person}{Nikhila Ravi}, \bibinfo{person}{Valentin Gabeur}, \bibinfo{person}{Yuan-Ting Hu}, \bibinfo{person}{Ronghang Hu}, \bibinfo{person}{Chaitanya Ryali}, \bibinfo{person}{Tengyu Ma}, \bibinfo{person}{Haitham Khedr}, \bibinfo{person}{Roman R{\"a}dle}, \bibinfo{person}{Chloe Rolland}, \bibinfo{person}{Laura Gustafson}, \bibinfo{person}{Eric Mintun}, \bibinfo{person}{Junting Pan}, \bibinfo{person}{Kalyan~Vasudev Alwala}, \bibinfo{person}{Nicolas Carion}, \bibinfo{person}{Chao-Yuan Wu}, \bibinfo{person}{Ross Girshick}, \bibinfo{person}{Piotr Doll{\'a}r}, {and} \bibinfo{person}{Christoph Feichtenhofer}.} \bibinfo{year}{2024}\natexlab{}.
\newblock \showarticletitle{SAM 2: Segment Anything in Images and Videos}.
\newblock \bibinfo{journal}{\emph{arXiv preprint arXiv:2408.00714}} (\bibinfo{year}{2024}).
\newblock
\urldef\tempurl%
\url{https://arxiv.org/abs/2408.00714}
\showURL{%
\tempurl}


\bibitem[Rombach et~al\mbox{.}(2022)]%
        {rombach2022high}
\bibfield{author}{\bibinfo{person}{Robin Rombach}, \bibinfo{person}{Andreas Blattmann}, \bibinfo{person}{Dominik Lorenz}, \bibinfo{person}{Patrick Esser}, {and} \bibinfo{person}{Bj{\"o}rn Ommer}.} \bibinfo{year}{2022}\natexlab{}.
\newblock \showarticletitle{High-resolution image synthesis with latent diffusion models}. In \bibinfo{booktitle}{\emph{Proceedings of the IEEE/CVF conference on computer vision and pattern recognition}}. \bibinfo{pages}{10684--10695}.
\newblock


\bibitem[Saharia et~al\mbox{.}(2022)]%
        {saharia2022photorealistic}
\bibfield{author}{\bibinfo{person}{Chitwan Saharia}, \bibinfo{person}{William Chan}, \bibinfo{person}{Saurabh Saxena}, \bibinfo{person}{Lala Li}, \bibinfo{person}{Jay Whang}, \bibinfo{person}{Emily~L Denton}, \bibinfo{person}{Kamyar Ghasemipour}, \bibinfo{person}{Raphael Gontijo~Lopes}, \bibinfo{person}{Burcu Karagol~Ayan}, \bibinfo{person}{Tim Salimans}, {et~al\mbox{.}}} \bibinfo{year}{2022}\natexlab{}.
\newblock \showarticletitle{Photorealistic text-to-image diffusion models with deep language understanding}.
\newblock \bibinfo{journal}{\emph{Advances in neural information processing systems}}  \bibinfo{volume}{35} (\bibinfo{year}{2022}), \bibinfo{pages}{36479--36494}.
\newblock


\bibitem[Schonberger and Frahm(2016)]%
        {schonberger2016structure}
\bibfield{author}{\bibinfo{person}{Johannes~L Schonberger} {and} \bibinfo{person}{Jan-Michael Frahm}.} \bibinfo{year}{2016}\natexlab{}.
\newblock \showarticletitle{Structure-from-motion revisited}. In \bibinfo{booktitle}{\emph{Proceedings of the IEEE conference on computer vision and pattern recognition}}. \bibinfo{pages}{4104--4113}.
\newblock


\bibitem[Shi et~al\mbox{.}(2025)]%
        {shi2025caring}
\bibfield{author}{\bibinfo{person}{Jingyu Shi}, \bibinfo{person}{Rahul Jain}, \bibinfo{person}{Seunggeun Chi}, \bibinfo{person}{Hyungjun Doh}, \bibinfo{person}{Hyung-gun Chi}, \bibinfo{person}{Alexander~J Quinn}, {and} \bibinfo{person}{Karthik Ramani}.} \bibinfo{year}{2025}\natexlab{}.
\newblock \showarticletitle{CARING-AI: Towards Authoring Context-aware Augmented Reality INstruction through Generative Artificial Intelligence}. In \bibinfo{booktitle}{\emph{Proceedings of the 2025 CHI Conference on Human Factors in Computing Systems}}. \bibinfo{pages}{1--23}.
\newblock


\bibitem[Sun et~al\mbox{.}(2024)]%
        {sun2024occfusion}
\bibfield{author}{\bibinfo{person}{Adam Sun}, \bibinfo{person}{Tiange Xiang}, \bibinfo{person}{Scott Delp}, \bibinfo{person}{Fei-Fei Li}, {and} \bibinfo{person}{Ehsan Adeli}.} \bibinfo{year}{2024}\natexlab{}.
\newblock \showarticletitle{Occfusion: Rendering occluded humans with generative diffusion priors}.
\newblock \bibinfo{journal}{\emph{Advances in Neural Information Processing Systems}}  \bibinfo{volume}{37} (\bibinfo{year}{2024}), \bibinfo{pages}{92184--92209}.
\newblock


\bibitem[Sun et~al\mbox{.}(2022)]%
        {sun2022direct}
\bibfield{author}{\bibinfo{person}{Cheng Sun}, \bibinfo{person}{Min Sun}, {and} \bibinfo{person}{Hwann-Tzong Chen}.} \bibinfo{year}{2022}\natexlab{}.
\newblock \showarticletitle{Direct voxel grid optimization: Super-fast convergence for radiance fields reconstruction}. In \bibinfo{booktitle}{\emph{Proceedings of the IEEE/CVF conference on computer vision and pattern recognition}}. \bibinfo{pages}{5459--5469}.
\newblock


\bibitem[Suvorov et~al\mbox{.}(2022)]%
        {suvorov2022resolution}
\bibfield{author}{\bibinfo{person}{Roman Suvorov}, \bibinfo{person}{Elizaveta Logacheva}, \bibinfo{person}{Anton Mashikhin}, \bibinfo{person}{Anastasia Remizova}, \bibinfo{person}{Arsenii Ashukha}, \bibinfo{person}{Aleksei Silvestrov}, \bibinfo{person}{Naejin Kong}, \bibinfo{person}{Harshith Goka}, \bibinfo{person}{Kiwoong Park}, {and} \bibinfo{person}{Victor Lempitsky}.} \bibinfo{year}{2022}\natexlab{}.
\newblock \showarticletitle{Resolution-robust large mask inpainting with fourier convolutions}. In \bibinfo{booktitle}{\emph{Proceedings of the IEEE/CVF winter conference on applications of computer vision}}. \bibinfo{pages}{2149--2159}.
\newblock


\bibitem[Teed and Deng(2020)]%
        {teed2020raft}
\bibfield{author}{\bibinfo{person}{Z. Teed} {and} \bibinfo{person}{J. Deng}.} \bibinfo{year}{2020}\natexlab{}.
\newblock \showarticletitle{RAFT: Recurrent All-Pairs Field Transforms for Optical Flow}. In \bibinfo{booktitle}{\emph{European Conference on Computer Vision (ECCV)}}. \bibinfo{pages}{402--419}.
\newblock


\bibitem[Wang et~al\mbox{.}(2024)]%
        {wang2024sea}
\bibfield{author}{\bibinfo{person}{Yihan Wang}, \bibinfo{person}{Lahav Lipson}, {and} \bibinfo{person}{Jia Deng}.} \bibinfo{year}{2024}\natexlab{}.
\newblock \showarticletitle{Sea-raft: Simple, efficient, accurate raft for optical flow}. In \bibinfo{booktitle}{\emph{European Conference on Computer Vision}}. Springer, \bibinfo{pages}{36--54}.
\newblock


\bibitem[Wang et~al\mbox{.}(2004)]%
        {wang2004image}
\bibfield{author}{\bibinfo{person}{Zhou Wang}, \bibinfo{person}{Alan~C Bovik}, \bibinfo{person}{Hamid~R Sheikh}, {and} \bibinfo{person}{Eero~P Simoncelli}.} \bibinfo{year}{2004}\natexlab{}.
\newblock \showarticletitle{Image quality assessment: from error visibility to structural similarity}.
\newblock \bibinfo{journal}{\emph{IEEE transactions on image processing}} \bibinfo{volume}{13}, \bibinfo{number}{4} (\bibinfo{year}{2004}), \bibinfo{pages}{600--612}.
\newblock


\bibitem[Wu et~al\mbox{.}(2022)]%
        {wu2022selfsupervised}
\bibfield{author}{\bibinfo{person}{J. Wu}, \bibinfo{person}{Z. Yang}, {and} \bibinfo{person}{H. Kim}.} \bibinfo{year}{2022}\natexlab{}.
\newblock \showarticletitle{Self-Supervised Amodal Reconstruction from Single Images}. In \bibinfo{booktitle}{\emph{European Conference on Computer Vision (ECCV)}}. \bibinfo{pages}{341--356}.
\newblock


\bibitem[Wu et~al\mbox{.}(2024)]%
        {wu2024thor}
\bibfield{author}{\bibinfo{person}{Qianyang Wu}, \bibinfo{person}{Ye Shi}, \bibinfo{person}{Xiaoshui Huang}, \bibinfo{person}{Jingyi Yu}, \bibinfo{person}{Lan Xu}, {and} \bibinfo{person}{Jingya Wang}.} \bibinfo{year}{2024}\natexlab{}.
\newblock \showarticletitle{Thor: Text to human-object interaction diffusion via relation intervention}.
\newblock \bibinfo{journal}{\emph{arXiv preprint arXiv:2403.11208}} (\bibinfo{year}{2024}).
\newblock


\bibitem[Xie et~al\mbox{.}(2024)]%
        {xie2023template_free}
\bibfield{author}{\bibinfo{person}{Xianghui Xie}, \bibinfo{person}{Bharat~Lal Bhatnagar}, \bibinfo{person}{Jan~Eric Lenssen}, {and} \bibinfo{person}{Gerard Pons-Moll}.} \bibinfo{year}{2024}\natexlab{}.
\newblock \showarticletitle{Template Free Reconstruction of Human-object Interaction with Procedural Interaction Generation}. In \bibinfo{booktitle}{\emph{IEEE Conference on Computer Vision and Pattern Recognition (CVPR)}}.
\newblock


\bibitem[Xu et~al\mbox{.}(2024)]%
        {xu2024amodal}
\bibfield{author}{\bibinfo{person}{Katherine Xu}, \bibinfo{person}{Lingzhi Zhang}, {and} \bibinfo{person}{Jianbo Shi}.} \bibinfo{year}{2024}\natexlab{}.
\newblock \showarticletitle{Amodal completion via progressive mixed context diffusion}. In \bibinfo{booktitle}{\emph{Proceedings of the IEEE/CVF Conference on Computer Vision and Pattern Recognition}}. \bibinfo{pages}{9099--9109}.
\newblock


\bibitem[Yang et~al\mbox{.}(2024)]%
        {yang2024gaussianobject}
\bibfield{author}{\bibinfo{person}{Chen Yang}, \bibinfo{person}{Sikuang Li}, \bibinfo{person}{Jiemin Fang}, \bibinfo{person}{Ruofan Liang}, \bibinfo{person}{Lingxi Xie}, \bibinfo{person}{Xiaopeng Zhang}, \bibinfo{person}{Wei Shen}, {and} \bibinfo{person}{Qi Tian}.} \bibinfo{year}{2024}\natexlab{}.
\newblock \showarticletitle{GaussianObject: High-Quality 3D Object Reconstruction from Four Views with Gaussian Splatting}.
\newblock \bibinfo{journal}{\emph{ACM Transactions on Graphics (TOG)}} \bibinfo{volume}{43}, \bibinfo{number}{6} (\bibinfo{year}{2024}), \bibinfo{pages}{1--13}.
\newblock


\bibitem[Yang et~al\mbox{.}(2023)]%
        {yang2023rerender}
\bibfield{author}{\bibinfo{person}{Shuai Yang}, \bibinfo{person}{Yifan Zhou}, \bibinfo{person}{Ziwei Liu}, {and} \bibinfo{person}{Chen~Change Loy}.} \bibinfo{year}{2023}\natexlab{}.
\newblock \showarticletitle{Rerender a video: Zero-shot text-guided video-to-video translation}. In \bibinfo{booktitle}{\emph{SIGGRAPH Asia 2023 Conference Papers}}. \bibinfo{pages}{1--11}.
\newblock


\bibitem[Zhang et~al\mbox{.}(2018)]%
        {zhang2018unreasonable}
\bibfield{author}{\bibinfo{person}{Richard Zhang}, \bibinfo{person}{Phillip Isola}, \bibinfo{person}{Alexei~A Efros}, \bibinfo{person}{Eli Shechtman}, {and} \bibinfo{person}{Oliver Wang}.} \bibinfo{year}{2018}\natexlab{}.
\newblock \showarticletitle{The unreasonable effectiveness of deep features as a perceptual metric}. In \bibinfo{booktitle}{\emph{Proceedings of the IEEE conference on computer vision and pattern recognition}}. \bibinfo{pages}{586--595}.
\newblock


\bibitem[Zhang et~al\mbox{.}(2024)]%
        {zhang2024avid}
\bibfield{author}{\bibinfo{person}{Zhixing Zhang}, \bibinfo{person}{Bichen Wu}, \bibinfo{person}{Xiaoyan Wang}, \bibinfo{person}{Yaqiao Luo}, \bibinfo{person}{Luxin Zhang}, \bibinfo{person}{Yinan Zhao}, \bibinfo{person}{Peter Vajda}, \bibinfo{person}{Dimitris Metaxas}, {and} \bibinfo{person}{Licheng Yu}.} \bibinfo{year}{2024}\natexlab{}.
\newblock \showarticletitle{Avid: Any-length video inpainting with diffusion model}. In \bibinfo{booktitle}{\emph{Proceedings of the IEEE/CVF Conference on Computer Vision and Pattern Recognition}}. \bibinfo{pages}{7162--7172}.
\newblock


\bibitem[Zhou et~al\mbox{.}(2023a)]%
        {zhou2023propainter}
\bibfield{author}{\bibinfo{person}{Shangchen Zhou}, \bibinfo{person}{Chongyi Li}, \bibinfo{person}{Kelvin~CK Chan}, {and} \bibinfo{person}{Chen~Change Loy}.} \bibinfo{year}{2023}\natexlab{a}.
\newblock \showarticletitle{Propainter: Improving propagation and transformer for video inpainting}. In \bibinfo{booktitle}{\emph{Proceedings of the IEEE/CVF international conference on computer vision}}. \bibinfo{pages}{10477--10486}.
\newblock


\bibitem[Zhou et~al\mbox{.}(2024)]%
        {zhou2024upscale}
\bibfield{author}{\bibinfo{person}{Shangchen Zhou}, \bibinfo{person}{Peiqing Yang}, \bibinfo{person}{Jianyi Wang}, \bibinfo{person}{Yihang Luo}, {and} \bibinfo{person}{Chen~Change Loy}.} \bibinfo{year}{2024}\natexlab{}.
\newblock \showarticletitle{Upscale-a-video: Temporal-consistent diffusion model for real-world video super-resolution}. In \bibinfo{booktitle}{\emph{Proceedings of the IEEE/CVF Conference on Computer Vision and Pattern Recognition}}. \bibinfo{pages}{2535--2545}.
\newblock


\bibitem[Zhou et~al\mbox{.}(2023b)]%
        {zhou2023amodal}
\bibfield{author}{\bibinfo{person}{X. Zhou}, \bibinfo{person}{Y. Li}, \bibinfo{person}{Z. Wang}, {and} \bibinfo{person}{T. Others}.} \bibinfo{year}{2023}\natexlab{b}.
\newblock \showarticletitle{Amodal Instance Segmentation with Transformers}. In \bibinfo{booktitle}{\emph{Proceedings of the IEEE/CVF Conference on Computer Vision and Pattern Recognition (CVPR)}}. \bibinfo{pages}{1234--1243}.
\newblock


\bibitem[Zhu et~al\mbox{.}(2024)]%
        {zhu2024fsgs}
\bibfield{author}{\bibinfo{person}{Zehao Zhu}, \bibinfo{person}{Zhiwen Fan}, \bibinfo{person}{Yifan Jiang}, {and} \bibinfo{person}{Zhangyang Wang}.} \bibinfo{year}{2024}\natexlab{}.
\newblock \showarticletitle{Fsgs: Real-time few-shot view synthesis using gaussian splatting}. In \bibinfo{booktitle}{\emph{European conference on computer vision}}. Springer, \bibinfo{pages}{145--163}.
\newblock


\end{thebibliography}

\clearpage
% \appendix
\twocolumn[
  \centering
  % First line, big bold title + line break + 1.5em of extra space
  {\fontsize{17pt}{20pt}\selectfont \textbf{Occlusion-Aware Temporally Consistent Amodal Completion for 3D Human-Object Interaction Reconstruction}}
  \vspace{2em}
  \\
  % Second line, smaller “Supplementary Material”
  {\huge Supplementary Material}
  \par\vspace{2em}  % finish the box and add a bit of vertical space
]
\setcounter{section}{0}
\renewcommand{\thesection}{\Alph{section}}
\renewcommand{\thesubsection}{\thesection.\arabic{subsection}}
\begin{figure*}
    \centering
    % \fbox{\makebox[0.90\linewidth][c]{\rule{0pt}{5cm} Ablation study for 2D temporal consistency}}
   \includegraphics[width=1\linewidth]{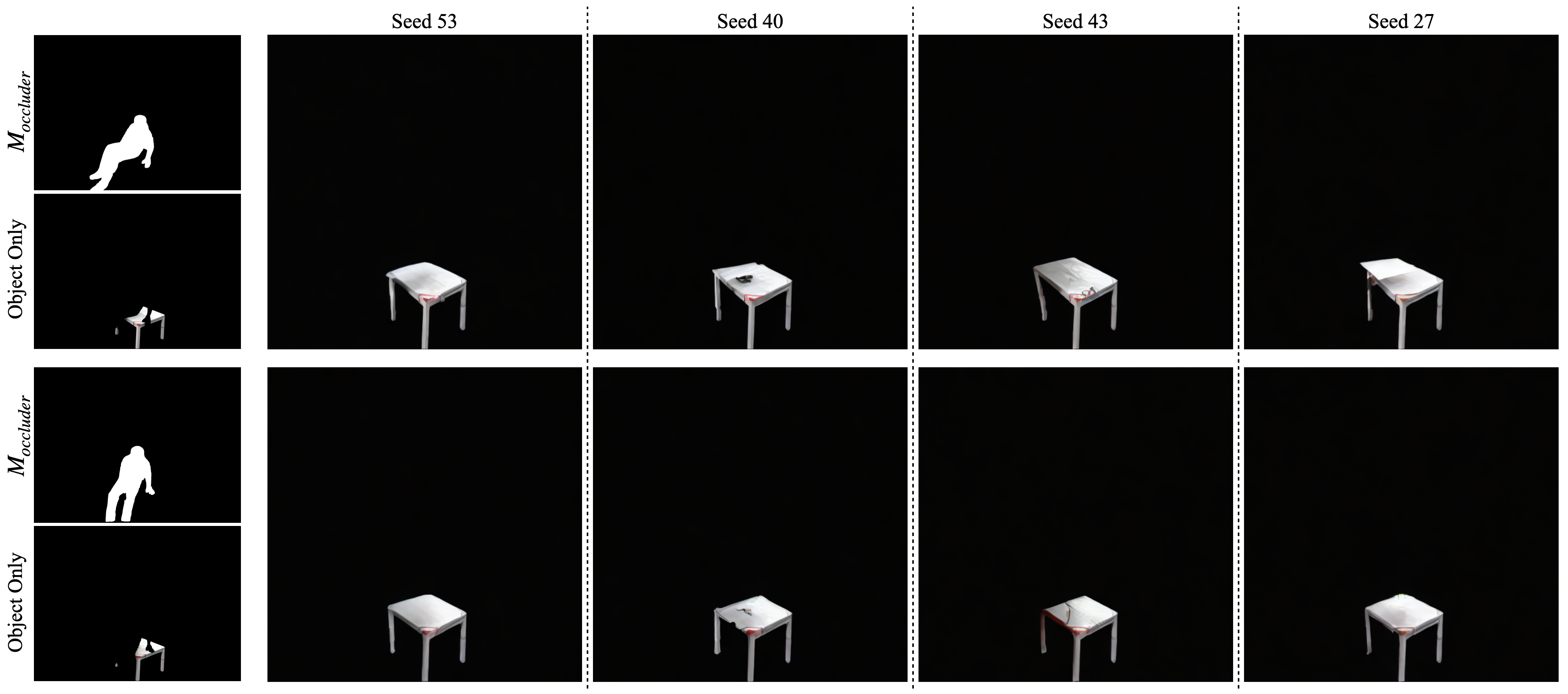}
   \vspace{-2.em}
    \caption{Fixed seed preliminary experiment for temporal consistency. Two frames are completed using four different random seeds to assess consistency.}
    \label{fig:hypo1_fixed_seed}
\end{figure*}

\begin{figure*}
    \centering
   \includegraphics[width=1\linewidth]{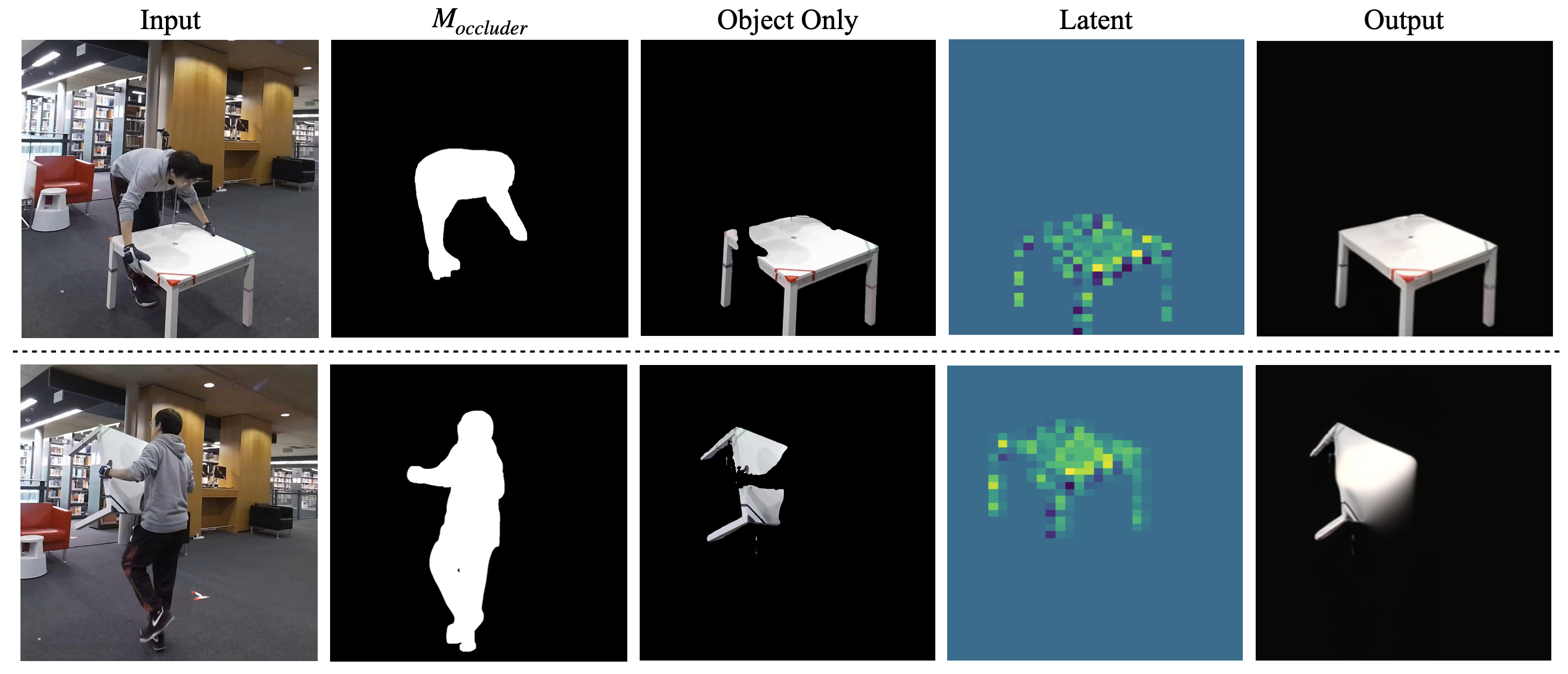}
   \vspace{-2.em}
    \caption{Preliminary experiment using latent shift for temporal consistency. The latent feature from the first frame is shifted to the second frame to evaluate whether temporal consistency can be maintained.}
    \label{fig:hypo2_latent_shift}
\end{figure*}

\begin{figure*}
    \centering
   \includegraphics[width=1\linewidth]{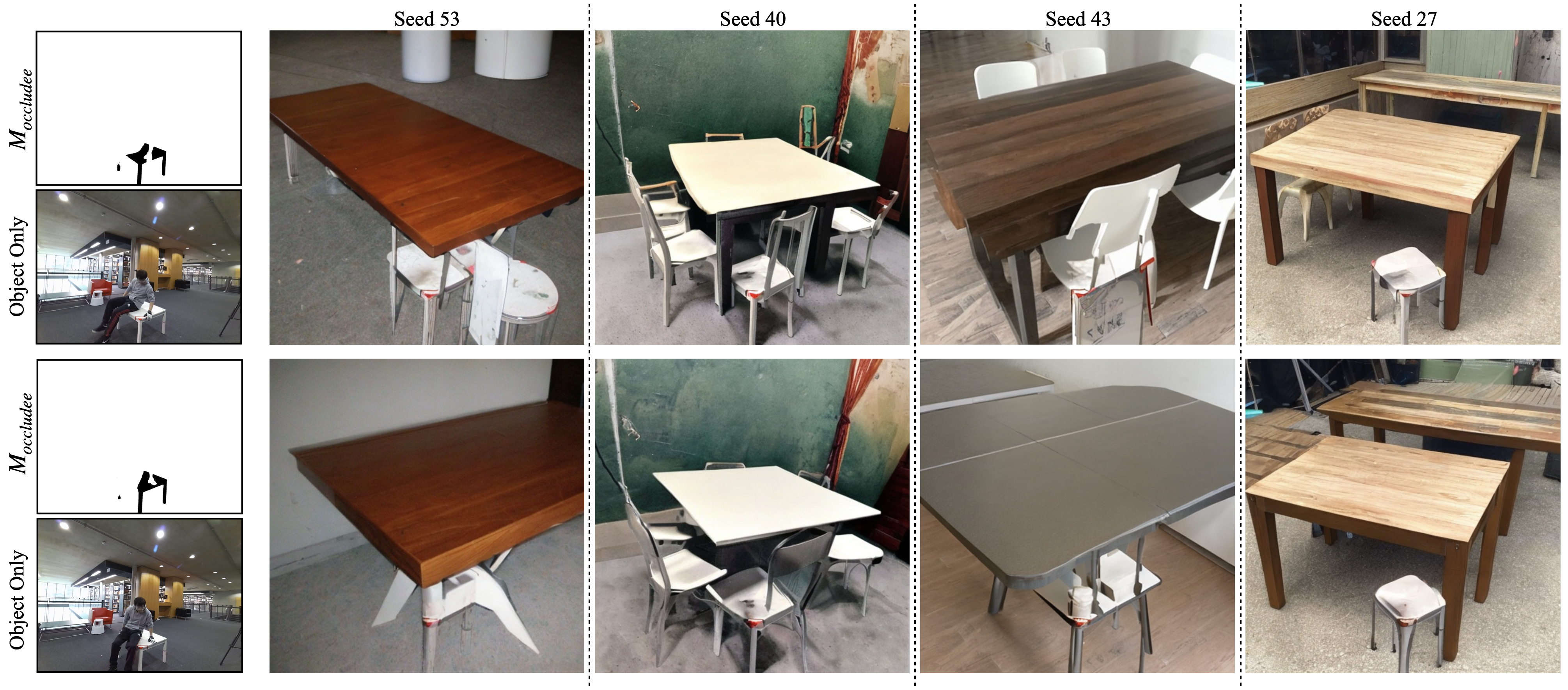}
   \vspace{-3.em}
    \caption{Preliminary experiment using a background mask for occlusion identification. The mask excludes only the unoccluded region of the target object.}
    \label{fig:hypo1_background_inpainting}
    \vspace{-1.em}
\end{figure*}

\section*{Overview}
This supplementary material introduces further details and experimental results of our paper, Temporally Consistent Amodal Completion for 3D Human-Object Interaction Reconstruction.

\begin{itemize}[leftmargin=*]
\item \Cref{sec:preliminary} elaborates on the preliminary experiments conducted prior to proposing our full pipeline.
\vspace{0.5em}

\item \Cref{sec:dataset} describes the datasets used, including BEHAVE~\cite{bhatnagar2022behave} and InterCap~\cite{huang2022intercap, huang2024intercap}.
\vspace{0.5em}

\item \Cref{sec:implementation_details} details the implementation of amodal completion and 3D reconstruction.
\vspace{0.5em}

\item \Cref{sec:evaluation} provides a detailed explanation of data selection process and the evaluation metrics for amodal completion, temporal consistency, and 3D reconstruction.
\vspace{0.5em}

\item \Cref{sec:additional_experiment} presents additional experiments on temporal consistency and occlusion identification.
\vspace{0.5em}

% \item \Cref{sec:discussion} provides additional discussion of our method.
% \vspace{0.5em}

\end{itemize}

\section{Preliminary Experiments}
\label{sec:preliminary}
Before proposing our full pipeline, we conducted preliminary experiments to explore temporal consistency and occlusion identification across frames. Specifically, we examined two strategies for achieving temporal consistency, fixing the random seed and applying a latent shift, and one strategy for occlusion identification, using a human mask and a background mask.

\paragraph{\textbf{Temporal Consistency with Fixed Random Seed.}}
~\Cref{fig:hypo1_fixed_seed} illustrates the effect of using a fixed random seed on temporal consistency. We generate two different frames with the same random seed, meaning that Stable Diffusion performs amodal completion from identical initial noise. As shown in the figure, the completion results lack consistency, even though the object remains static during the human-object interaction in both frames. This experiment demonstrates that fixing the random seed alone is insufficient to ensure temporal consistency. 

\paragraph{\textbf{Temporal Consistency with Latent Shift.}}
Since fixing the random seed was ineffective in maintaining temporal consistency, we hypothesized that leveraging latent features extracted from the original frame could help preserve consistency across frames. To isolate the effect of the latent representation, we applied a mask to retain only the target object's region.

However, due to the dynamic nature of human-object interaction, the target object moves across frames. To address this, we hypothesized that spatially shifting the latent features could mitigate temporal inconsistency. As shown in \Cref{fig:hypo2_latent_shift}, we applied the latent from the first frame to subsequent frames by aligning it with the target object's new position, without accounting for rotation. While the shifted latent features improved temporal consistency to some extent, the results suggest that spatial shifting alone—without adjusting for the object's rotation—is insufficient to fully preserve temporal consistency and accurately complete occluded regions.

\paragraph{\textbf{Occlusion Identification with Human Mask and Background Mask.}}
According to ~\Cref{fig:hypo1_fixed_seed}, applying a human mask does not reliably capture occluded regions. Moreover, using the entire background as a mask, excluding only the unoccluded region of the target object, results in an entirely new object that aligns only with the text prompt, as illustrated in ~\Cref{fig:hypo1_background_inpainting}. As a result, the completed areas often deviate from the true object shape, indicating that the outputs are not temporally consistent and geometrically accurate.

\section{Dataset}
\label{sec:dataset}
\textbf{BEHAVE~\cite{bhatnagar2022behave}.}  
The BEHAVE dataset consists of 321 RGB-D video sequences of indoor human–object interactions, recorded using four Kinect cameras.  
The test set includes 3 human subjects interacting with 20 different objects.  
For evaluation, we select 18 object sequences—excluding “keyboard” and “basketball” due to missing ground-truth pose annotations in the 30 FPS version.  
Additionally, we select 3 sequences, one for each subject, to evaluate human performance.  
In total, our pipeline is applied to approximately 27,000 frames across all selected sequences.

\noindent \textbf{InterCap~\cite{huang2022intercap, huang2024intercap}.}  
The InterCap dataset comprises 223 RGB-D videos of human–object interactions, captured from 6 camera viewpoints and involving 10 objects and 10 human subjects.  
We extract 10 representative videos that collectively cover all 10 object categories used in the dataset.

\section{Implementation Details}
\label{sec:implementation_details}

\subsection{Baselines}

We compare our approach against several state-of-the-art baselines for amodal completion and video inpainting. The selected methods represent diverse design philosophies and serve to evaluate different aspects of our method, including spatial plausibility and temporal consistency.

\paragraph{Pix2Gestalt~\cite{ozguroglu2024pix2gestalt}.}
This method performs amodal completion by hallucinating occluded object parts via semantic segmentation and image generation. Although designed for static images, it is widely adopted in amodal completion literature and serves as a strong baseline for appearance-level plausibility under occlusion.

\paragraph{Stable Diffusion Inpainting (SD Inpainting)~\cite{rombach2022high}.}
A popular diffusion-based inpainting method that generates high-quality content guided by masked regions. We adopt this model as a zero-shot inpainting baseline to assess the generative quality of completed regions. However, it lacks temporal modeling, making it susceptible to frame-wise inconsistency in videos.

\paragraph{LaMa~\cite{suvorov2022resolution}.}
LaMa is a fast, high-resolution image inpainting model that leverages fast Fourier convolutions. Its strong performance in image-level inpainting tasks makes it a competitive baseline, especially for spatial reconstruction fidelity. Like SD Inpainting, it does not consider temporal coherence.

\paragraph{VDT~\cite{lu2023vdt}.}
VDT is a video diffusion transformer trained for temporally consistent video inpainting. It leverages a space-time attention mechanism and is specifically designed for video scenarios. This method serves as a strong baseline for evaluating the temporal stability of inpainted sequences.

\subsection{Amodal Completion.}

We use Stable Diffusion Inpainting~\cite{rombach2022high} with a guidance scale of 6.0, keeping all other hyperparameters consistent with the default configuration. For Bidirectional Temporal Feature (BTF) Warping, we adopt SeaRaft~\cite{wang2024sea} for optical flow estimation and follow its official implementation settings. The window size \(n\) is empirically set to 7 based on performance comparison, as shown in~\Cref{tab:BTF_window}; smaller window sizes (\(n\) = 1, 3) result in inferior performance across all evaluation metrics.

\vspace{-0.5em}
\subsection{3D reconstruction.}

For object reconstruction, we use 30{,}000 training iterations and set the weighting parameter \(\lambda\) = 0.2 for the photometric loss \(\mathcal{L}_{photo}\), which combines L1 and SSIM terms. All other hyperparameters follow the settings of GS-Pose~\cite{mei2024gs2pose} for objects and GaussianAvatar~\cite{hu2024gaussianavatar} for humans.

Since the 30 FPS object poses and SMPL parameters provided in the BEHAVE dataset~\cite{bhatnagar2022behave} are not temporally aligned or spatially accurate, we observed noticeable misalignment between the rendered mesh and the inpainted object masks generated by SAM2~\cite{ravi2024sam2}. To mitigate this issue during 3D Gaussian Splatting~\cite{kerbl3Dgaussians} training, we use the provided object masks aligend with 6-DoF object poses instead of our predicted masks for object reconstruction. This assumes the inpainting process yields complete and reliable textures. Nevertheless, for evaluation purposes, we evaluate reconstruction performance using outputs generated solely by our full pipeline.

\begin{table}[t]
  \centering
  % Add commands to shrink table content
  \small 
  \setlength{\tabcolsep}{4pt} % Default is 6pt

  \begin{minipage}[t]{0.49\columnwidth}
    \centering
    \begin{tabular}{@{}lcc@{}}
      \toprule
      \textbf{Object} & \textbf{Size} & \textbf{Volume} \\
      \midrule
      tablesquare      & 557.56 & 1.68 \\
      chairwood        & 396.51 & 0.68 \\
      chairblack       & 294.38 & 1.56 \\
      tablesmall       & 233.93 & 0.42 \\
      yogaball         & 196.26 & 8.27 \\
      monitor          & 150.48 & 1.00 \\
      boxlarge         & 139.68 & 6.35 \\
      plasticcontainer & 132.87 & 0.82 \\
      yogamat          & 125.55 & 3.24 \\
      \bottomrule
    \end{tabular}
  \end{minipage}
  \hfill % Adds horizontal space
  \begin{minipage}[t]{0.49\columnwidth}
    \centering
    \begin{tabular}{@{}lcc@{}}
      \toprule
      \textbf{Object} & \textbf{Size} & \textbf{Volume} \\
      \midrule
      boxlong          & 92.17  & 0.96 \\
      stool            & 87.29  & 2.64 \\
      backpack         & 83.14  & 2.98 \\
      suitcase         & 67.32  & 3.65 \\
      boxmedium        & 40.44  & 2.96 \\
      boxsmall         & 26.73  & 1.29 \\
      trashbin         & 30.45  & 1.75 \\
      \textbf{toolbox}& \textbf{13.34}  & \textbf{0.52} \\
      \textbf{boxtiny}& \textbf{6.96}   & \textbf{0.27} \\
      \bottomrule
    \end{tabular}
  \end{minipage}
  
  \caption{3D bounding box size and volume for each object in BEHAVE [4], sorted by descending size. \textit{boxtiny} and \textit{toolbox} have the smallest spatial extent and are excluded from 3D reconstruction experiments.}
  \label{tab:object_stats_sorted}
  \vspace{-2.0em}
\end{table}

\section{Evaluation}
\label{sec:evaluation}

\paragraph{\textbf{Data Selection.}}  
To ensure rigorous and fair evaluation, we carefully curated samples from the BEHAVE~\cite{bhatnagar2022behave} and InterCap~\cite{huang2022intercap, huang2024intercap} datasets based on the level of object occlusion.  
Specifically, we include only frames in which the object occlusion ratio falls between 15\% and 70\%.  
Frames with minimal occlusion provide limited challenge for amodal completion, while those with severe occlusion lack sufficient visual cues for meaningful reconstruction.  
This strict filtering balances task difficulty, ensures evaluation stability, and prevents inflation of dataset size with low-signal or overly ambiguous cases.

\paragraph{\textbf{Amodal Completion.}}  
To evaluate the quality of amodal completion, we use two complementary metrics: CLIP Score~\cite{radford2021learning} and Intersection over Union (IoU). 
The CLIP score quantifies the semantic alignment between each inpainted image and a corresponding category-level textual prompt using CLIP's vision–language embedding space. This metric has been widely adopted in various tasks~\cite{saharia2022photorealistic, poole2022dreamfusion, lee2025editsplat} without ground truth data to evaluate.
This allows us to assess whether the completed region semantically resembles the intended object class.  
To reduce the impact of unrelated background pixels, we compute CLIP scores within tight bounding boxes centered on the reconstructed objects.

For spatial accuracy, we report the IoU between the predicted amodal masks and the ground-truth masks.  
Since our model outputs RGB images rather than masks, we employ SAM2~\cite{ravi2024sam2} to segment the inpainted regions and generate amodal predictions.  
IoU is then computed as the ratio of the intersection and union areas between the predicted and ground-truth object masks.

\paragraph{\textbf{Temporal Consistency.}}  
We assess temporal consistency using two metrics: the Temporal Consistency score (TC score)~\cite{esser2023structure} and Flow Warping Error~\cite{lai2018learning}.

The TC score is designed to measure perceptual smoothness across time.  
Specifically, we extract CLIP image embeddings from each frame of the output video and compute the average cosine similarity between embeddings of consecutive frames.  
A higher TC score implies better visual consistency over time, as perceptually similar frames yield higher similarity in CLIP space.

Flow Warping Error captures temporal misalignment at the pixel level.  
We first estimate optical flow between pairs of consecutive ground-truth object masks using SEA-RAFT~\cite{wang2024sea}.  
Using the flow, the previously inpainted frame is warped to align with the current frame.  
The pixel-wise discrepancy between the warped frame and the actual current frame is then measured as the Flow Warping Error.  
A lower Flow Warping Error indicates higher temporal coherence and fewer frame-to-frame inconsistencies.

\begin{figure}
    \centering
    % \fbox{\makebox[0.90\linewidth][c]{\rule{0pt}{5cm} Ablation study for 2D temporal consistency}}
   \includegraphics[width=1\linewidth]{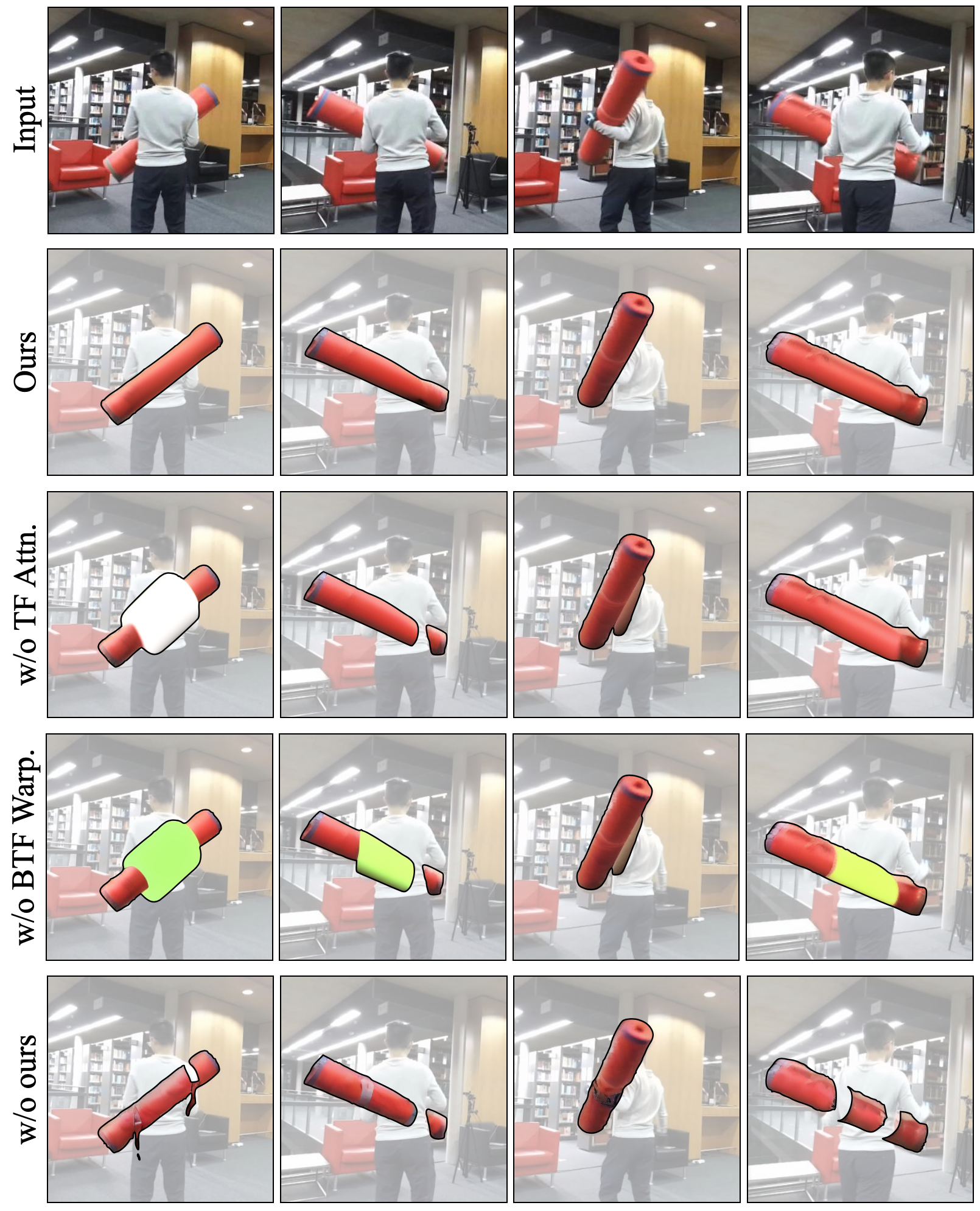}
   \vspace{-2.em}
    \caption{Qualitative Ablation of Our methods.}
    \label{fig:2d_ablation}
\end{figure}

\paragraph{\textbf{3D reconstruction.}}

We evaluate 3D reconstruction on 16 out of the 18 objects used in our amodal completion and temporal consistency experiments. Following our setup, we sample video frames at 1 FPS, using only 15–47 frames per sequence. Two objects (“box tiny” and “tool box”) are excluded from evaluation due to their extremely small projected area and volume, which result in poor reconstruction quality (see \Cref{tab:object_stats_sorted} for object statistics).

Since our method reconstructs only the object or the human the background—we evaluate reconstruction quality using masked variants of standard metrics: Peak Signal-to-Noise Ratio (PSNR-M), Structural Similarity Index Measure (SSIM-M)\cite{wang2004image}, and Learned Perceptual Image Patch Similarity (LPIPS-M)\cite{zhang2018unreasonable}. Following the evaluation protocol in~\cite{nazarczuk2024aim}, we compute these metrics within tight axis-aligned bounding boxes surrounding the reconstructed human and object regions, rather than over the entire frame. To minimize background influence and better focus on the target regions, we crop around the reconstructed masks using a tight object mask with a small scale margin (×1.2), ensuring a highly localized evaluation.

This cropped evaluation is crucial because baseline methods often produce geometrically inconsistent reconstructions that fail to preserve the spatial extent of the original subject. As a result, evaluating over the full image would unfairly penalize such methods due to misalignment with the ground truth. By focusing the evaluation within localized bounding boxes, we reduce the influence of irrelevant background pixels and obtain a more faithful assessment of perceptual and geometric reconstruction quality.

\section{Additional Experiments}
\label{sec:additional_experiment}
\paragraph{\textbf{Ablation Study on Temporal Consistency.}}

Figure~\ref{fig:2d_ablation} shows qualitative results comparing different ablations. Our method maintains consistent appearances across consecutive frames, while ablated variants suffer from noticeable temporal inconsistencies. This highlights the effectiveness of our temporal consistency strategy.

\paragraph{\textbf{Effect of Temporal Window Size.}}
To evaluate the impact of temporal context, we conduct an ablation study on the window size $n$ used in the Bidirectional Temporal Feature (BTF) Warping module, as shown in~\Cref{tab:BTF_window}. We compare three settings: $n = 1$, $n = 3$, and $n = 7$ (our default). Results show that increasing the window size leads to consistent improvements across all metrics. Specifically, our setting ($n = 7$) achieves the highest IoU (61.23), CLIP score (27.81), and temporal consistency (97.18), while also yielding the lowest warping error (5.69). This demonstrates that incorporating a broader temporal context enables more accurate feature alignment and robust reconstruction, particularly in the presence of complex motion and occlusion.

\section{Potential Applications}
Our current work demonstrates novel pose synthesis by borrowing human and object poses from different videos. However, this approach can be significantly extended by integrating with text-to-motion and text-to-HOI models such as \cite{chi2024m2d2m, wu2024thor, shi2025caring}. By adapting human and object poses from the outputs of text prompts, our model can achieve greater flexibility in HOI animation.

\begin{table}
  \centering
  \renewcommand{\arraystretch}{1.3}
  \begin{tabular}{c c c c c}
    \toprule
    \textbf{Window Size \(n\)} & {IoU \(\uparrow\)} & {CLIP \(\uparrow\)} & {\parbox{1.2cm}{\centering Warp-err\\(x$10^{-3}$)} \(\downarrow\)} & {TC \(\uparrow\)} \\
    % \textbf{Feature Warping} & \textbf{Cross Attention} & {IoU} & {CLIP} & {Warp-err (x$10^{-3}$)} & {TC Score} \\
    \hline\hline
    {1} & {60.84} & {27.80} & {5.73} & {97.17} \\
    {3} & {60.75} & {27.80} & {5.70} & {97.17} \\
    {7 (\textbf{Ours})} & \textbf{61.23} & \textbf{27.81} & \textbf{5.69} & \textbf{97.18} \\
    \hline
  \end{tabular}
  % \vspace{0.5em}
  \caption{Effect of window size \(n\) in Bidirectional Temporal Feature (BTF) Warping. Using a larger window (\(n = 7\)) provides improved alignment and better performance across IoU, CLIP score, warp error, and temporal consistency (TC), compared to smaller window sizes.}
  \vspace{-2.em}
  \label{tab:BTF_window}
\end{table}

\end{document}